\documentclass{article}
\usepackage{iclr2027_conference,times}

\usepackage{amsmath,amsfonts,bm}

\def\eqref#1{equation~\ref{#1}}

\def\1{\bm{1}}

\DeclareMathAlphabet{\mathsfit}{\encodingdefault}{\sfdefault}{m}{sl}
\SetMathAlphabet{\mathsfit}{bold}{\encodingdefault}{\sfdefault}{bx}{n}

\newcommand{\R}{\mathbb{R}}

\usepackage[utf8]{inputenc}
\usepackage[T1]{fontenc}
\usepackage{hyperref}
\usepackage{url}
\usepackage{booktabs}
\usepackage{amsfonts}
\usepackage{amsmath,amssymb}
\usepackage{amssymb}
\usepackage{nicefrac}
\usepackage{microtype}
\usepackage{xcolor}
\usepackage{graphicx}
\usepackage{algorithm}
\usepackage{multirow}
\usepackage{arydshln}
\usepackage{enumitem}
\usepackage[utf8]{inputenc}
\usepackage[T1]{fontenc}
\usepackage{hyperref}
\usepackage{url}
\usepackage{booktabs}
\usepackage{amsfonts}
\usepackage{nicefrac}
\usepackage{microtype}
\usepackage{xcolor}
\usepackage{amsmath,amssymb,amsthm}
\usepackage{algorithm}
\usepackage{algpseudocode}
\usepackage{graphicx} 
\usepackage{subcaption}

\newcommand{\heng}[1]{#1}
\DeclareMathOperator{\Gr}{Gr}

\newcommand{\haotian}[1]{#1}

\newtheorem{theorem}{Theorem}
\newtheorem{lemma}[theorem]{Lemma}

\newtheorem{corollary}{Corollary}

\theoremstyle{definition}
\newtheorem{assumption}{Assumption}

\theoremstyle{remark}
\newtheorem{remark}{Remark}

\newcounter{obscounter}
\renewcommand{\theobscounter}{\arabic{obscounter}}
\makeatletter
\newcommand{\obspara}[1]{\refstepcounter{obscounter}\paragraph{Observation \theobscounter: #1}}
\makeatother

\title{FLAME: Federated Laplace-Aligned Mixture of Experts for Personalized LLM Fine-Tuning}

\author{%
  Haotian Xiang$^{1}$ \quad
  Heng Zhang$^{1}$ \quad
  Ping Xu$^{2}$ \quad
  Qin Lu$^{1}$\thanks{Corresponding author.} \\[0.6em]
  $^{1}$University of Georgia \\
  $^{2}$Georgia State University
}

\iclrfinalcopy

\begin{document}

\maketitle
\lhead{Preprint}

\begin{abstract}
Personalized federated fine-tuning of large language models calls for adaptation to heterogeneous client tasks together with reliable uncertainty estimates under limited local data. Existing personalized federated low-rank adaptation (LoRA) methods expose only a single level of cross-client locality, average the LoRA factors $A$ and $B$ identically despite their structural asymmetry, and return {\it point} estimate adapters whose predictions remain poorly calibrated. To address these limitations, we propose {\bf FLAME}, a {\bf F}ederated {\bf L}aplace-{\bf A}ligned {\bf M}ixture of {\bf E}xperts whose three design choices are each grounded in the geometry of the learned parameter space. To capture hierarchical locality across client adaptations, FLAME jointly routes among self, intra-cluster, and cross-cluster experts, with the trainable individual expert active in every forward pass. To resolve the LoRA basis ambiguity that Euclidean aggregation introduces, FLAME aligns the input-side factor $A$ on the Grassmann manifold and keeps the output-side factor $B$ in Euclidean space, where the entry-level task-specific signals live. For {\it calibrated} per-client predictive distributions, FLAME fits a local post-hoc Kronecker-factored Laplace posterior over each client’s individual expert, leaving training and communication untouched. We further establish a non-asymptotic convergence guarantee for FLAME and show that the Grassmannian aggregator tightens the aggregation-drift more than Euclidean averaging.
Experimental results across $12$-client MCQA classification benchmark and $8$-client FLAN NLG benchmarks spanning IID, cross-dataset, and label-skew heterogeneity show that FLAME consistently improves both accuracy and calibration (ECE, NLL) over all baselines.
\end{abstract}

\vspace{-1em}
\section{Introduction}
\label{sec:intro}
\vspace{-0.5em}

Modern data sources such as hospitals, banks, and edge devices generate text at scale, yet privacy regulations and institutional policies prevent these data from being centralized for training \citep{mcmahan2017communication}. Federated learning (FL) addresses this constraint by allowing collaborative training across distributed clients while keeping raw data local \citep{mcmahan2017communication, tan2022towards}. Deploying FL on top of large language models (LLMs) \citep{kenton2019bert, brown2020language, touvron2023llama} is a natural next step, since adapting LLMs to specialized domains typically relies on fine-tuning with task-specific data \citep{minaee2024large} that is itself federated by nature. Full fine-tuning of an LLM on every client is, however, computationally prohibitive and incurs per-round communication on the order of the entire backbone, which has motivated the federated adoption of parameter-efficient fine-tuning (PEFT) methods \citep{ding2023parameter}. Among PEFT methods, low-rank adaptation (LoRA) \citep{hu2022lora, li2026low} has become the de facto standard: LoRA inserts trainable low-rank matrices $A$ and $B$ alongside frozen pretrained weights and updates far fewer parameters per adapted layer, reducing both local compute and per-round communication by orders of magnitude relative to full fine-tuning, which makes it the natural adapter to use under FL constraints \citep{zhang2023building, kuang2024federatedscope, ye2024openfedllm, fan2023fate}.

This federated LoRA setup introduces a complication absent from centralized fine-tuning: in realistic LLM deployments, clients often hold data from genuinely different downstream tasks rather than mere label shifts of a shared task, and a single globally averaged LoRA degrades per-client performance under such task heterogeneity. \emph{Personalized} federated LoRA addresses this by allowing each client to retain some adapter state, with FedDPA \citep{long2024dual} blending a global LoRA with a local LoRA at inference, FedALT \citep{fedalt} freezing the rest-of-world (RoW) LoRA during local training and routing it against a per-client individual LoRA, and FedLEASE \citep{fedlease} replacing the single RoW adapter with a clustered mixture of experts. Despite their different routing structures, the server-side aggregator of these methods remains FedAvg-style and averages $A$ and $B$ identically in Euclidean space.

A closer look at the trained federated LoRA parameter space shows that the server-side aggregator is poorly matched to the geometry it operates on. Three structural properties go unexploited by current designs, each raising a research question for personalized federated LoRA. We measure all three properties directly on the trained federated LoRA parameter space in Section~\ref{sec:asym}, and preview them here as \emph{locality structure}, \emph{gauge ambiguity and asymmetry}, and \emph{calibration gap}.
\emph{(1) Locality structure.} Existing personalized routers expose only a single level of cross-client locality, whether binary \citep{long2024dual, fedalt} or per-cluster \citep{fedlease}, while the trained $A$-factors carry non-redundant locality at both intra-cluster and cross-cluster scales, with row-space structure shared among intra-cluster peers \citep{collins2021exploiting} but absent across cluster boundaries. How should an architecture expose both scales jointly with each client's self-adaptation, and how should the cluster count be set without manual tuning?
\emph{(2) Gauge ambiguity and asymmetry.} The shared aggregator is doubly mismatched to LoRA. The factorization \citep{hu2022lora} is non-unique up to an invertible change of basis on the rank-$r$ space, so independently trained clients converge to different factor pairs and Euclidean averaging drifts in this null space \citep{sun2024improving, florg2026}. The two factors also carry the signal asymmetrically \citep{tian2024hydra, zhang2023adalora}, and the asymmetry \emph{inverts} across scales. Shared row-space structure across intra-cluster peers lives on $A$, whereas task identity is encoded in the entries of $B$. What is the right aggregation operator on each factor, and which factor should drive the clustering?
\emph{(3) Calibration gap.} Architecture and aggregation aside, per-client adaptation on limited local data is the regime where calibrated predictive uncertainty matters most for selective prediction and downstream decision-making \citep{mackay1992practical, wilson2020bayesian}, yet uncertainty quantification has remained largely unexplored for personalized federated LLMs, with existing federated UQ work \citep{achituve2021personalized, kassab2022federated} targeting sub-LLM-scale models. Centralized Bayesian LoRA methods do exist \citep{daxberger2021laplace, ritter2018scalable}, though they either modify the training loop \citep{wang2024blob, yang2023bayesian} or add per-step memory cost \citep{wang2023lora, tian2024variational} that does not scale across federated clients. Can a Bayesian recipe deliver calibrated uncertainty for personalized federated LLMs while leaving training or communication intact?

We propose \textbf{FLAME} (\textbf{F}ederated \textbf{L}aplace-\textbf{A}ligned \textbf{M}ixture of \textbf{E}xperts), a personalized federated LoRA framework with three components, each matched to one of the properties above. For locality, FLAME assigns each client a trainable individual adapter together with a set of frozen experts at the intra-cluster and cross-cluster scales, and routes among them while the trainable individual adapter takes part in every forward pass. 
To address the gauge ambiguity on $A$ and the $A/B$ asymmetry jointly, FLAME applies a \emph{selective Grassmannian aggregator} that aggregates $A$ through a gauge-invariant Fr\'{e}chet mean on the rank-$r$
Grassmannian and $B$ retain the Euclidean mean where the entry-level task signal lives. 
\heng{For calibration, FLAME fits a post-hoc
Kronecker-factored (KFAC) Laplace posterior over each
client's individual adapter, storing a pair of Kronecker
factors for each LoRA factor and leaving training and
per-round communication untouched.} Figure~\ref{fig:method_compare} contrasts FLAME with representative baselines along locality, aggregation geometry, and uncertainty treatment.

Our contributions are summarized as follows.
\begin{itemize}[nosep,leftmargin=*]
\item We identify three structural properties of the trained federated LoRA parameter space that existing personalized federated LoRA designs leave unexploited: cross-client locality at three non-redundant scales, an $A/B$ asymmetry that inverts between subspace and entry scales, and a calibration gap that persists across routing architectures.
\item We propose FLAME with three matched components: (i) a three-granularity (self / intra-cluster / cross-cluster) expert hierarchy with guaranteed per-token participation of the individual adapter; (ii) a \emph{selective Grassmannian aggregator} resolving LoRA's gauge ambiguity on $A$ via a Fr\'{e}chet mean on the rank-$r$ Grassmannian while keeping Euclidean averaging on $B$; (iii) a per-client post-hoc KFAC Laplace posterior fitted without extra communication.
\item We provide a non-asymptotic convergence analysis for FLAME's training in which each design component maps to an identifiable term of the bound, with the Grassmannian aggregator strictly tightening the aggregation-drift term over Euclidean averaging.
\item We show on a $12$-client MCQA classification benchmark and three $8$-client FLAN NLG benchmarks under varying heterogeneity that FLAME outperforms competitive baselines in every setting, with the calibration improvement unmatched by scaling the LoRA rank of a point-estimate baseline.
\end{itemize}

\vspace{-0.5em}\section{Preliminaries and Motivations}
\label{sec:prelim}
\vspace{-0.05cm}
\subsection{LoRA and Federated LoRA}
Consider a federated setting with $N$ clients. Each client $n$ ($n=1,\ldots,N$) possesses a local dataset $\mathcal{D}_n$ that may originate from a different downstream task. All clients share a frozen pretrained LLM whose weight at every adapted linear layer is denoted $W^0\in\R^{d_{\rm out}\times d_{\rm in}}$, where $d_{\rm in}$ and $d_{\rm out}$ are the input and output dimensions. LoRA \citep{hu2022lora} fine-tunes a low-rank residual $\Delta W=BA$ with $A\in\R^{r\times d_{\rm in}}$, $B\in\R^{d_{\rm out}\times r}$, and $r\ll\min(d_{\rm in},d_{\rm out})$, so that the adapted weight is $W = W^0 + BA$. The input-side factor $A$ is initialized by Kaiming sampling, while the output-side factor $B$ is initialized to zero, so that the residual at the start of training is exactly zero. The personalized federated fine-tuning objective is to learn a personalized adapted model $W_n = W^0 + B_n A_n$ that fits $\mathcal{D}_n$,
\begin{equation}
\min_{\{(A_n,B_n)\}_{n=1}^N} \frac{1}{N}\sum_{n=1}^N \mathcal{L}_n\bigl( A_n, B_n; \mathcal{D}_n\bigr),
\label{eq:objective}
\end{equation}
where $\mathcal{L}_n$ is the local task loss for client $n$. The standard FedAvg \citep{mcmahan2017communication} aggregator returns
\begin{equation}
\bar{A}=\frac{1}{N}\sum_{n=1}^{N} A_n,
\qquad
\bar{B}=\frac{1}{N}\sum_{n=1}^{N} B_n.
\label{eq:fedavg}
\end{equation}
The aggregator in \eqref{eq:fedavg} has two structural issues. Averaging $A$ and $B$ independently does not commute with the matrix product, so $\bar{B}\bar{A}\neq\tfrac{1}{N}\sum_n B_n A_n$ in general, which introduces a bilinear aggregation error that is absent when full-weight FedAvg is used \citep{florg2026}. \heng{For a peer set $\mathcal{P}$, directly averaging
the products $B_mA_m$ can yield an update of rank up
to $|\mathcal{P}|r$, which may no longer admit the
rank-$r$ expert parameterization in
\eqref{eq:moe_forward}.}
The factorization $\Delta W_n = B_n A_n$ is also non-unique up to a change of basis on the rank-$r$ latent space: for any invertible $R\in\mathrm{GL}(r)$, where $\mathrm{GL}(r)$ is the group of $r\times r$ invertible real matrices, the pair $(R A_n, B_n R^{-1})$ yields the same effective update $(B_n R^{-1})(R A_n) = B_n A_n$. Two clients can therefore represent the same $\Delta W_n$ with completely different $(A,B)$ pairs related by some such $R$, and Euclidean averaging of either factor across clients drifts in this gauge null space, a phenomenon we refer to as \emph{decomposition drift}.
These two issues, together with the asymmetric initialization (Kaiming for $A$, zero for $B$) that drives qualitatively different update trajectories during local training, indicate that $A$ and $B$ are not interchangeable under aggregation, and that any aggregator applying the same operator to both inherits these errors. To decide whether and how the two factors should be aggregated differently, we examine the structure that emerges in each from the trained federated LoRA parameter space, leading to the three observations of Section~\ref{sec:asym} that motivate FLAME's design.

\begin{figure}[t]
\centering
\includegraphics[width=0.80\linewidth]{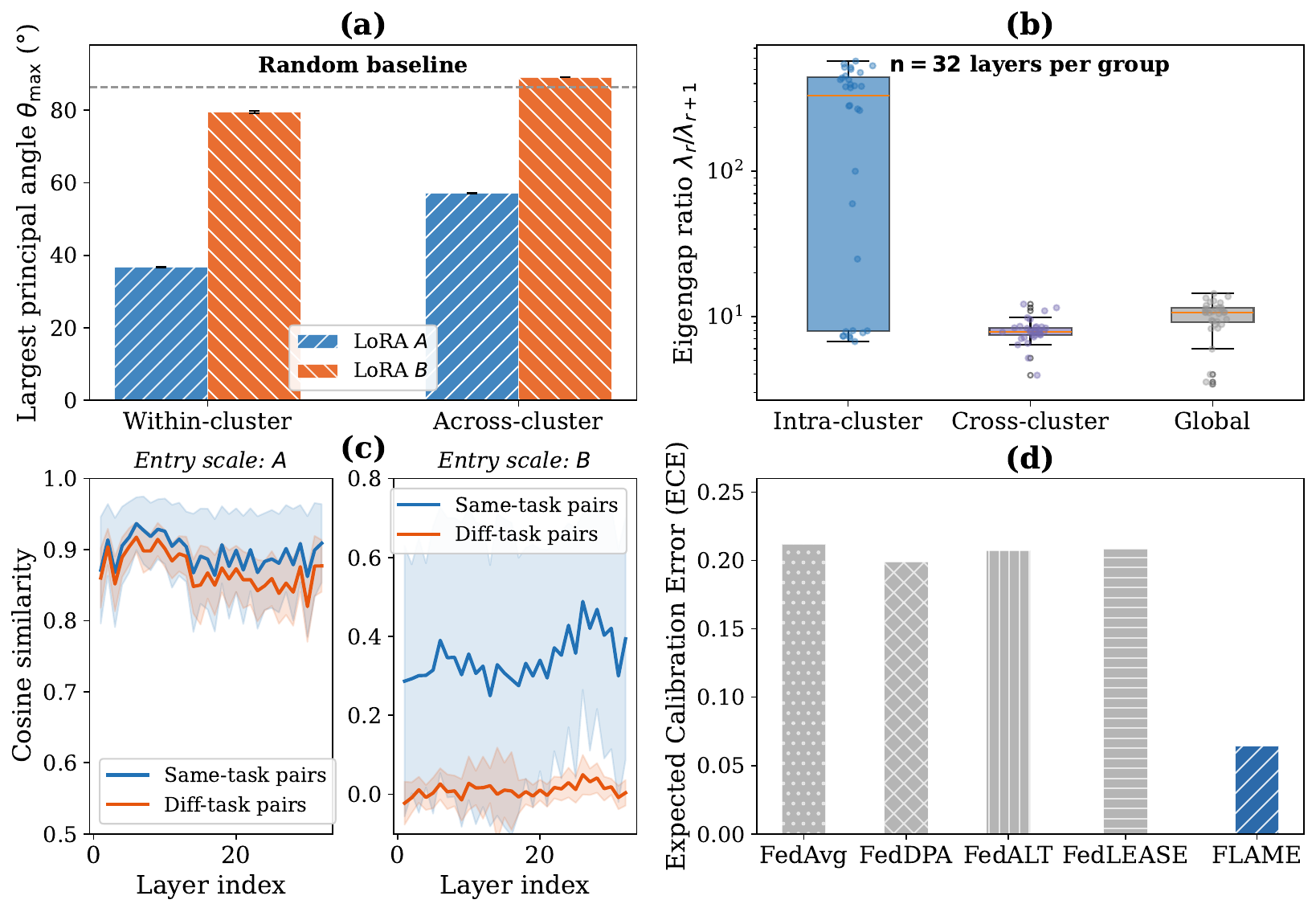}
\vspace{-0.2cm}
\caption{Three observations on the trained federated LoRA parameter space. \emph{(a)} Largest principal angle between client $A$/$B$ subspaces at intra- and cross-cluster granularities, against the random rank-$r$ baseline (dashed). \emph{(b)} Eigengap ratio $\lambda_r/\lambda_{r+1}$ of the average $A$-subspace projector at intra-cluster, cross-cluster, and global peer sets ($n{=}32$ layers per group); larger ratio $\rightarrow$ better-separated rank-$r$ subspace. \emph{(c)} Per-layer entry-scale cosine similarity on $A$ and $B$ between same-task and different-task client pairs. \emph{(d)} Aggregate ECE of four federated LoRA approaches against FLAME's full pipeline.}
\label{fig:observations}
\vspace{-1em}
\end{figure}

\vspace{-0.6em}\subsection{Observations}
\label{sec:asym}
We report three observations on the trained federated LoRA parameter space that motivate FLAME's design. The studies use the MCQA-12c benchmark (defined in Section~\ref{sec:exp-setup}), with $N=12$ clients spanning $4$ downstream tasks at $3$ clients per task and LoRA rank $r=8$ at every adapted linear layer. Each client first runs one round of local-only fine-tuning of its own $(A_n,B_n)$ on $\mathcal{D}_n$ without aggregation, and the trained factors are then analyzed.

\vspace{-0.8em}\obspara{$A$ carries shared structure at the subspace scale only.}\label{obs:A}
We measure two properties of the trained $A$-factors against the random rank-$r$ baseline. The largest principal angle between client $A$-subspaces (Figure~\ref{fig:observations}(a)) shows tight alignment at the intra-cluster scale and substantially weaker alignment across distinct task clusters. The eigengap ratio $\lambda_r/\lambda_{r+1}$ of the average $A$-subspace projector (Figure~\ref{fig:observations}(b)), where a larger ratio indicates a better-separated rank-$r$ subspace, is well-resolved at the intra-cluster scale and collapses by one to two orders of magnitude at both cross-cluster and global peer sets. The per-layer cosine similarity of $A$ at the entry scale (Figure~\ref{fig:observations}(c), $A$ panel) shows no separation between same-task and different-task client pairs at any layer. The signal on $A$ therefore lives in its row space, and only at the intra-cluster scale does that row space line up across peers. A Grassmannian aggregator on $A$ at the cluster level matches this geometry directly.

\vspace{-0.8em} \obspara{$B$ carries identifying signal at the entry scale only.}\label{obs:B}
We apply the same two measurements to $B$. At the subspace scale (Figure~\ref{fig:observations}(a), $B$ bars), the largest principal angle between client $B$-subspaces sits at or beyond the random rank-$r$ baseline at both intra-cluster and cross-cluster peers, leaving the row space of $B$ statistically indistinguishable from a random subspace and offering nothing for an alignment step to exploit. At the entry scale (Figure~\ref{fig:observations}(c), $B$ panel), the per-layer cosine similarity cleanly separates same-task from different-task client pairs at every adapted layer, so the task signal on $B$'s lives in its entries. Comparing these results with Observations~\ref{obs:A}, the federated signal sits on $A$ and $B$ at inverted scales. On $A$, it lives in the row-space and only among intra-cluster peers. On $B$, it lives directly in the entries. A single aggregator applied to both factors at the same scale therefore reads one signal as the other's noise, which motivates a selective aggregator that aligns $A$ on the Grassmann manifold while averaging $B$ in Euclidean space.

\vspace{-0.8em} \obspara{the calibration gap is invariant to the routing approach.}\label{obs:calib}
Figure~\ref{fig:observations}(d) reports the aggregate expected calibration error (ECE) of four federated LoRA approaches spanning the personalization spectrum, from a single global LoRA (FedAvg) to a clustered mixture of experts (FedLEASE). 
The four approaches reach similar ECE levels despite substantial differences in routing architecture, and the component ablation in Section~\ref{sec:exp-ablation} (Table~\ref{tab:abl_component}) shows that FLAME's own architecture without any Bayesian treatment also lies in the same range, while FLAME's full pipeline brings ECE substantially lower. Each approach returns a maximum-a-posteriori (MAP) adapter whose output is a single softmax over data already seen during training. The calibration gap is, therefore, a property of point estimation rather than of any particular routing layer, and closing it calls for a Bayesian treatment that is orthogonal to the routing.

\begin{figure*}[t]
\centering
\includegraphics[width=0.90\textwidth]{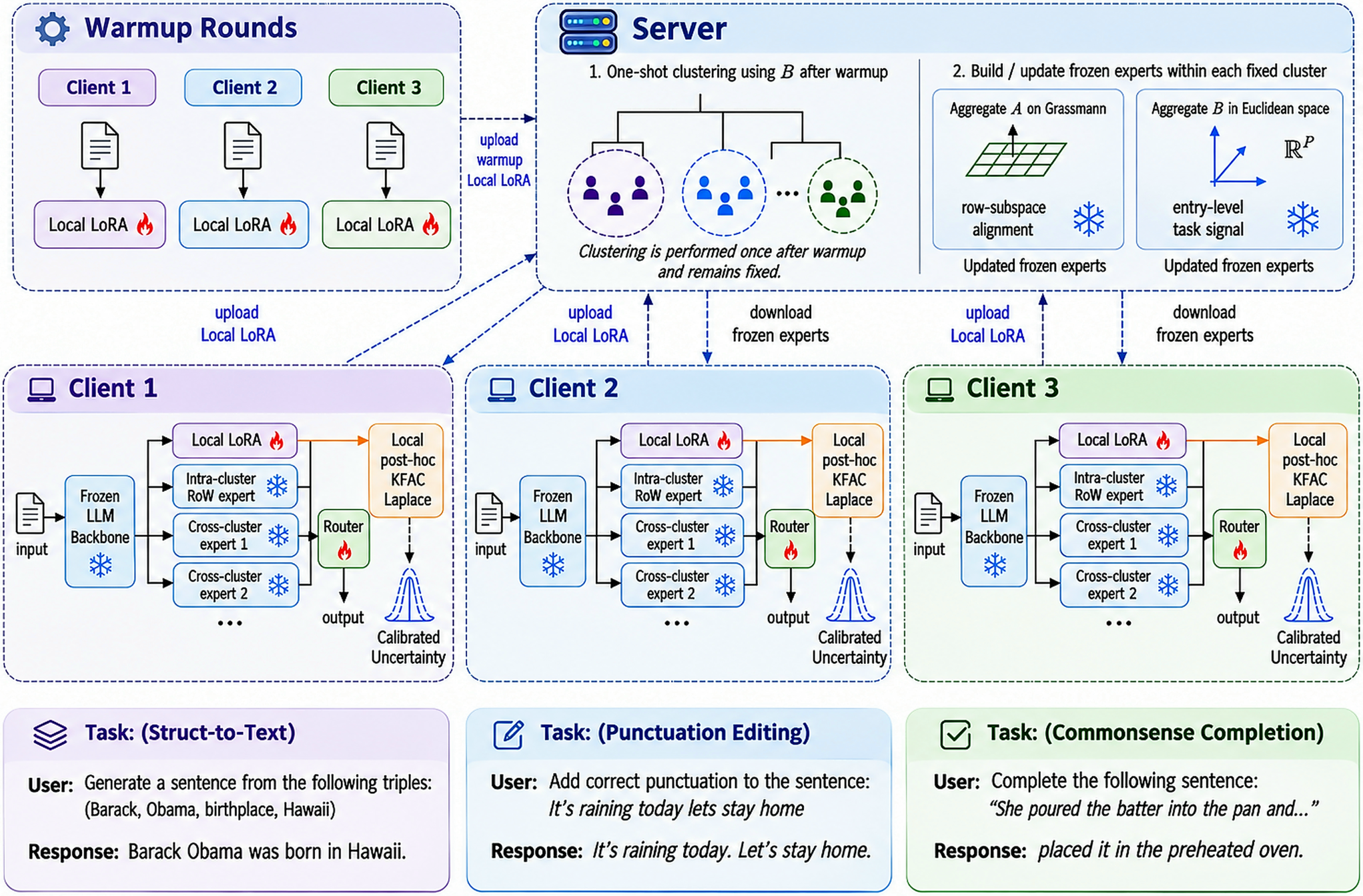}
\vspace{-0.3cm}
\caption{
Overview of FLAME. 
Clients first perform a local warmup and upload their local LoRA factors to the server. 
After warmup, the server performs one-shot clustering using the output-side factor $B$, then builds frozen experts within each fixed cluster by aggregating $A$ on the Grassmann manifold and $B$ in Euclidean space. 
During personalized federated training, each client combines its trainable local LoRA with intra-cluster and cross-cluster frozen experts through a learned router. 
After training converges, each client fits a local post-hoc KFAC Laplace approximation on its individual expert to produce a predictive mean and calibrated uncertainty.
}
\label{fig:flame_overview}
\vspace{-0.5cm}
\end{figure*}
\section{FLAME: Federated Laplace-Aligned Mixture of Experts}
\label{sec:method}

The three observations of Section~\ref{sec:asym} identify three properties of the trained federated LoRA parameter space that no existing personalized federated LoRA method jointly addresses. We next present FLAME, a framework that introduces three components matched to these three properties. The end-to-end FLAME pipeline is presented in Figure~\ref{fig:flame_overview}.

\subsection{Personalized clustered mixture of experts}
\label{sec:method-backbone}

The federated training in FLAME consists of three stages. The first stage initiates a brief local warmup to produce task-specific representations from which the server can infer cluster structure. The second stage is a one-shot clustering step at the server, and the third stage involves clustered mixture-of-experts (MoE) training for the remaining rounds. 
\heng{Starting from the same initialization of $A$ and
$B$, every client trains its own LoRA $(A_n,B_n)$ on
$\mathcal{D}_n$ for $T_w$ warmup rounds, after which
the warmed-up factors are uploaded once to the server.} The server then forms $M$ clusters by hierarchical agglomerative clustering \citep{fedlease} on the pairwise distance matrix
\begin{equation}
D_{n,m} \;=\; \frac{1}{L}\sum_{\ell=1}^{L}\Bigl(1 - \cos\bigl(\mathrm{vec}(B_n^{(\ell)}),\mathrm{vec}(B_m^{(\ell)})\bigr)\Bigr),
\label{eq:cluster_dist}
\end{equation}
where $\mathrm{vec}(\cdot)$ flattens a matrix into a vector, $\cos(\cdot,\cdot)$ is the cosine, $L$ is the number of adapted linear layers, and $B_n^{(\ell)}$ is client $n$'s output-side factor at layer $\ell$. The number of clusters $M$ is selected by silhouette score over $\{2,\ldots,M_{\max}\}$ with average linkage, where $M_{\max}$ is a hyperparameter upper bound on the cluster count, and the result is the partition $\{{\cal C}_j\}_{j=1}^M$ in which ${\cal C}_j$ collects all clients assigned to the $j$-th cluster. Clustering is performed on $B$ rather than $A$, indicated by Observation~\ref{obs:B} that $B$ carries entry-level task identity whereas $A$'s entries are dominated by the shared Kaiming initialization at this stage and provide no task-discriminative signal.

After clustering, every client $n$ in cluster ${\cal C}_j$ is assigned $M+1$ LoRA experts spanning three disjoint granularities, matched one to one with the locality levels of Observation~\ref{obs:A}. The \emph{individual expert} at granularity~0 holds the client's own trainable parameters $(A_n,B_n)$. The \emph{intra-cluster RoW expert} at granularity~1 is the frozen leave-one-out aggregate $(\bar{A}_{\mathcal{C}_j\setminus\{n\}},\bar{B}_{\mathcal{C}_j\setminus\{n\}})$ \citep{fedalt} over the client's cluster peers, with the aggregation operator described in Section~\ref{sec:method-grass}. The $M-1$ \emph{cross-cluster experts} at granularity~2 are frozen aggregates $(\bar{A}_{\mathcal{C}_{j'}},\bar{B}_{\mathcal{C}_{j'}})$ for each peer cluster $j'\neq j$ over all of its members. The leave-one-out construction at granularity~1 ensures that the frozen signal received by client $n$ is genuinely out-of-self and therefore complements the trainable individual expert without redundancy.

We adopt the enlarged top-$k$ router of \citep{fedlease}, where the individual expert $(A_n,B_n)$ occupies multiple output positions while each frozen expert occupies a single position. With $k=M{+}1$, this construction guarantees that the individual expert participates in every forward pass. Specifically, $\phi_n$ is a per-layer trainable linear layer, $\mathcal{T}(x)$ denotes the index set of the top $M{+}1$ outputs of $\phi_n(x)$, and $\pi(i)\in\{0,1,\ldots,M\}$ maps each output position $i$ to its wired expert, with $\pi(i)=0$ for the individual. The LoRA update at every adapted layer is
\begin{equation}
\Delta W(x) \;=\; \sum_{i\in\mathcal{T}(x)} g_i(x)\, B^{(\pi(i))} A^{(\pi(i))} x,
\qquad
g_i(x) \;=\; \frac{\exp(\phi_n(x)_i)}{\sum_{j\in\mathcal{T}(x)} \exp(\phi_n(x)_j)},
\label{eq:moe_forward}
\end{equation}
where $(A^{(0)},B^{(0)})=(A_n,B_n)$ is the trainable individual expert and $(A^{(e)},B^{(e)})$ for $e\in\{1,\ldots,M\}$ are frozen aggregates. The total per-token gate weight assigned to the individual expert,
\begin{equation}
g_I(x) \;:=\; \sum_{i\in\mathcal{T}(x):\,\pi(i)=0} g_i(x),
\label{eq:gate_I}
\end{equation}
admits a structural lower bound $g_I(x)\ge g_{i^\star}(x)>0$ at every $x$, where $i^\star$ is any selected position wired to the individual expert. Throughout this clustered training stage, gradients flow only into $(A_n,B_n)$ and the router $\phi_n$, while the frozen experts and $W^0$ remain unchanged; at the end of every round, client $n$ uploads only $(A_n,B_n)$ to the server, and the per-layer router $\phi_n$ stays on the client. \heng{For clients with multiple local tasks, the
token-level router allows tokens from different tasks
to draw on different experts, including the
cross-cluster experts.}

\subsection{Grassmannian alignment of $A$ at the server side}
\label{sec:method-grass}

With the MoE-based clustering mechanism, we now specify how the frozen experts in \eqref{eq:moe_forward} are constructed from the LoRA factors uploaded by clients. To avoid the decomposition drift observed in Observation~\ref{obs:A} for the input-side factor, we aggregate 
$A$ at every level via the chordal-distance Fr\'{e}chet mean on the Grassmann manifold $\Gr(r,d_{\rm in})$, while keeping the standard Euclidean mean for $B$ at every level, in accordance with the asymmetry of Observation~\ref{obs:B}.

Let $\mathcal{P}\subseteq\{1,\ldots,N\}\setminus\{n\}$ denote a peer set whose factors are to be aggregated: $\mathcal{P}=\mathcal{C}_j\setminus\{n\}$ for the intra-cluster RoW expert of client $n\in \mathcal{C}_j$, and $\mathcal{P}=\mathcal{C}_{j'}$ for some $j'\neq j$ for cross-cluster expert $e\in\{2,\ldots,M\}$. \heng{Each peer set is restricted to clients that
upload their factors in the current round.
An absent client retains its individual and previously
received frozen experts, and receives the current
frozen experts upon returning.} \heng{For a single-client cluster, the intra-cluster
peer set is empty. The client therefore routes over
its individual and cross-cluster experts, with the
individual expert guaranteed to participate in
every forward pass.} The Grassmannian aggregator is computed in three steps. First, every peer's $A_m$ is orthonormalized via thin QR decomposition of $A_m^\top$ to give $Q_m=\mathrm{orth}(A_m)\in\R^{r\times d_{\rm in}}$ whose rows form an orthonormal basis of $\mathrm{rowspan}(A_m)$, and the average orthogonal projector onto these row spaces is computed as
\begin{equation}
\bar{P}_\mathcal{P} \;=\; \frac{1}{|\mathcal{P}|}\sum_{m\in\mathcal{P}} Q_m^\top Q_m,
\label{eq:proj_avg}
\end{equation}
where each $Q_m^\top Q_m\in\R^{d_{\rm in}\times d_{\rm in}}$ is invariant under the gauge transformation $A_m\mapsto R A_m$, so $\bar{P}_\mathcal{P}$ depends only on the row spaces themselves rather than on any specific basis. Second, the consensus basis is read off as the top-$r$ eigenvectors of $\bar{P}_\mathcal{P}$,
\begin{equation}
\bar{Q}_\mathcal{P} \;=\; \bigl[\,\mathrm{top}\text{-}r\text{-}\mathrm{eigvecs}(\bar{P}_\mathcal{P})\,\bigr]^\top \;\in\;\R^{r\times d_{\rm in}},
\label{eq:consensus}
\end{equation}
which is the chordal-distance Fr\'{e}chet mean on $\Gr(r,d_{\rm in})$ in closed form \citep{absil2004riemannian}. Third, each peer's coefficients are projected onto the consensus basis and averaged,
\begin{equation}
\bar{A}_\mathcal{P} \;=\; \biggl(\frac{1}{|\mathcal{P}|}\sum_{m\in\mathcal{P}} A_m \bar{Q}_\mathcal{P}^{\top}\biggr)\bar{Q}_\mathcal{P}.
\label{eq:coeff_avg}
\end{equation}
\heng{By Lemma~\ref{lem:gauge}, the projectors
$Q_m^\top Q_m$, their average $\bar{P}_\mathcal{P}$,
and the consensus projector are invariant under
client-specific transformations $A_m\mapsto R_m A_m$
with $R_m\in\mathrm{GL}(r)$.
Thus, the consensus row space onto which
\eqref{eq:coeff_avg} projects the aggregate is
independent of each client's choice of basis.}
The output-side factor uses the standard Euclidean mean
\begin{equation}
\bar{B}_\mathcal{P} \;=\; \frac{1}{|\mathcal{P}|}\sum_{m\in\mathcal{P}} B_m.
\label{eq:b_mean}
\end{equation}
\heng{If either sum in
Eqs.~\eqref{eq:coeff_avg}--\eqref{eq:b_mean}
vanishes for a nonempty peer set, the corresponding
frozen expert produces a zero update in ~\eqref{eq:moe_forward}, while the individual
expert remains active in every forward pass.} The frozen experts in \eqref{eq:moe_forward} are then $(A^{(1)},B^{(1)})=(\bar{A}_{\mathcal{C}_j\setminus\{n\}},\bar{B}_{\mathcal{C}_j\setminus\{n\}})$ for the intra-cluster RoW and $(A^{(e)},B^{(e)})=(\bar{A}_{\mathcal{C}_{j_e}},\bar{B}_{\mathcal{C}_{j_e}})$ for cross-cluster expert $e\geq 2$, where $j_e$ enumerates the other clusters.

The eigengap measurement in Figure~\ref{fig:observations}(b) constrains where this aggregator may be applied. The intra-cluster eigengap is well-resolved, while choosing the peer set $\mathcal{P}$ to span multiple clusters collapses the eigengap by one to two orders of magnitude and leaves $\bar{Q}_\mathcal{P}$ underdetermined. FLAME therefore restricts the Grassmannian aggregator to single-cluster peer sets at every level of the hierarchy: the intra-cluster RoW expert uses $\mathcal{P}=\mathcal{C}_j\setminus\{n\}$, which is a leave-one-out subset of one cluster, and each cross-cluster expert uses $\mathcal{P}=\mathcal{C}_{j'}$ for one peer cluster $j'\neq j$, rather than letting $\mathcal{P}$ span multiple clusters. 
\heng{Both peer sets sit in the high-eigengap regime of
Figure~\ref{fig:observations}(b), providing well-resolved
consensus subspaces for frozen-expert construction.
Moreover, the factors averaged in
\eqref{eq:coeff_avg} and \eqref{eq:b_mean} originate
from a common initialization of $A$ and $B$, with
each peer set restricted to a single cluster.}

\subsection{Local post-hoc KFAC Laplace at local agents}
\label{sec:method-laplace}

\heng{The MoE mechanism in
Section~\ref{sec:method-backbone} returns a
maximum-a-posteriori (MAP) adapter.
As shown in Observation~\ref{obs:calib}, point
estimation leaves a calibration gap across routing
architectures. To obtain calibrated predictive
distributions, each client locally fits a post-hoc
Kronecker-factored Laplace posterior over its
individual-expert factors $(A_n,B_n)$
\citep{daxberger2021laplace,martens2015kfac,
ritter2018scalable,yang2024laplace}.
Following Laplace-LoRA, we treat $A_n$ and $B_n$
as two separate linear layers, each with its own
pair of Kronecker factors.
The Laplace stage leaves training and per-round
communication unchanged.}

For an adapted layer on client $n$, let
$F\in\{A,B\}$ index the two individual-expert factors.
Let $a_{n,t}^{F}$ and $\delta_{n,t}^{F}$ denote
the input activation and output gradient of factor
$F$, respectively, at sample $t$ of the local
dataset $\mathcal{D}_n$, with
$T_n=|\mathcal{D}_n|$.
For each factor, the KFAC Gram matrices are

\begin{equation}
\heng{
\Omega_n^{F}
=
\frac{1}{T_n}\sum_{t=1}^{T_n}
a_{n,t}^{F}(a_{n,t}^{F})^\top,
\qquad
\Lambda_n^{F}
=
\frac{1}{T_n}\sum_{t=1}^{T_n}
\delta_{n,t}^{F}(\delta_{n,t}^{F})^\top.
}
\label{eq:kfac_factors}
\end{equation}

For $A_n$, the input and output dimensions
are $d_{\rm in}$ and $r$, respectively, while for
$B_n$ they are $r$ and $d_{\rm out}$.
Both pairs of Kronecker factors are accumulated
in one forward-backward sweep over $\mathcal{D}_n$,
with the frozen experts and trained router fixed
and without parameter updates.
\heng{The likelihood curvature is summed over the
$T_n$ local samples, giving the KFAC approximation
$T_n(\Omega_n^{F}\otimes\Lambda_n^{F})$
in column-wise vectorization
\citep{ritter2018scalable,yang2024laplace}. The sample-count scaling allows the posterior
uncertainty to reflect the amount of local data.
An isotropic Gaussian prior with precision
$\alpha_n>0$ is used, where $\alpha_n$ is selected
per client by maximizing the Kronecker-factored
log marginal likelihood
$\log p(\mathcal{D}_n;\alpha_n)$.}

\heng{The Laplace expansion is centered at the trained
factors $A_{n,\mathrm{MAP}}$ and $B_{n,\mathrm{MAP}}$
in the same coordinates used during local optimization.
For each factor $F\in\{A,B\}$, the KFAC posterior
precision is approximated by
$P_n^F\approx T_n(\Omega_n^F\otimes\Lambda_n^F)
+\alpha_n I$.
We draw $S$ posterior samples for each factor:}

\begin{equation}
\mathrm{vec}(F_{n,s})
\sim
\mathcal{N}\left(
\mathrm{vec}(F_{n,\mathrm{MAP}}),
(P_n^F)^{-1}
\right),
\quad
F\in\{A,B\},\quad s=1,\ldots,S.
\label{eq:posterior_sample}
\end{equation}

Each posterior sample forms a low-rank
individual-expert update
$\Delta W_{n,s}=B_{n,s}A_{n,s}$,
preserving the rank-$r$ LoRA parameterization.
The sampled individual experts replace their MAP
counterparts at each adapted layer, while the frozen
experts and trained router remain fixed.
The posterior-predictive distribution is

\begin{equation}
p_n(y\mid x)
\approx
\frac{1}{S}\sum_{s=1}^{S}
\left[
\mathrm{softmax}\bigl(f(x;A_{n,s},B_{n,s})\bigr)
\right]_y.
\label{eq:bma}
\end{equation}

where $f(x;A_{n,s},B_{n,s})$ denotes the model
logits with the sampled individual-expert factors
at every adapted layer.
All posterior quantities are computed locally on
$\mathcal{D}_n$, introducing no additional
communication.

\vspace{-0.5em}\paragraph{Remark (Why Laplace fits FLAME?)}

The recipe of
\eqref{eq:kfac_factors}-\eqref{eq:bma}
relies on three preconditions, all satisfied by
FLAME's mechanism.

\heng{\emph{MAP validity}:
the Laplace expansion is centered at the trained
individual-expert factors $(A_n,B_n)$, in the
same coordinates used during local optimization
with the frozen experts fixed in the forward pass
\citep{daxberger2021laplace}.}  \heng{\emph{Fisher well-definedness}:
the KFAC factors in \eqref{eq:kfac_factors}
accumulate curvature separately for $A_n$ and $B_n$,
the parameters over which the posterior is defined.}  \emph{BMA leverage}:
the lower bound on $g_I(x)$ guaranteed by the router
\eqref{eq:moe_forward} keeps the individual expert's
posterior contribution bounded away from zero at
every token, ensuring non-vanishing Bayesian model
averaging weight \citep{wilson2020bayesian};
FedALT's unconstrained two-expert mixer admits no
such bound, and $g_I(x)$ can collapse on tokens
dominated by the RoW expert.
We confirm this empirically in
Section~\ref{sec:exp-ablation}
(Table~\ref{tab:lap_universal}):
equipping FedALT with the same post-hoc Laplace
stage still leaves a calibration gap to FLAME.

\section{Theoretical Analysis}
\label{sec:theory}

We analyze the convergence of FLAME's clustered MoE training stage (Section~\ref{sec:method-backbone}), in which the trainable individual expert $(A_n,B_n)$ is the only parameter group that receives a local gradient signal. Let $K=TE$ be the total number of local SGD steps across $T$ rounds with $E$ steps per round, $\mathcal{L}_n^t$ the round-$t$ local loss on client $n$, and $\Delta_T=\sum_{t=1}^{T-1}\|\bar{\mathbf{E}}^{t+1}-\bar{\mathbf{E}}^t\|$  the cumulative inter-round drift of the frozen-expert configuration. \heng{Let $\mathcal{C}_S$ denote the Lipschitz
constant of the local per-sample loss $F_n$ with
respect to the frozen-expert configuration
(Assumption~\ref{ass:freeze-smooth}), namely,
$|F_n(\theta;\xi;\bar{\mathbf{E}},G_n)
-F_n(\theta;\xi;\bar{\mathbf{E}}',G_n)|
\leq \mathcal{C}_S
\|\bar{\mathbf{E}}-\bar{\mathbf{E}}'\|$.} Here, we only state  the main result, with the full descriptions of assumptions, notation, and proofs delegated to  Appendix~\ref{sec:convergence}; 

\noindent{\bf Theorem 1} ({\bf Convergence of FLAME).} {\it Under the assumptions in Appendix~\ref{sec:conv-assumptions}, the local SGD iterates produced by FLAME with constant step size $\eta\in(0,1/L_\Phi]$, for every client $n$, satisfy
\begin{equation}
\label{eq:thm-body}
\frac{1}{K}\sum_{k=0}^{K-1}\mathbb{E}\bigl\|\nabla\mathcal{L}_n^{t(k)}(\theta_n^{(k)})\bigr\|^2
\;\leq\;
\frac{2\bigl(\mathcal{L}_n^1(\theta_n^{(0)})-\mathcal{L}_n^\star\bigr)}{\eta K}
\;+\;L_\Phi\,\eta\,G_\Phi^2
\;+\;\frac{2\mathcal{C}_S\Delta_T}{\eta K}.
\end{equation}
With $\eta=\min\!\bigl(1/L_\Phi,\sqrt{2D_0/(L_\Phi G_\Phi^2K)}\bigr)$ and $D_0=\mathcal{L}_n^1(\theta_n^{(0)})-\mathcal{L}_n^\star+\mathcal{C}_S\Delta_T$, the right-hand side decays at rate $\mathcal{O}\!\bigl(K^{-1/2}\bigr)$. } 

Each term on the right side of \eqref{eq:thm-body} can be traced to a component of FLAME. The smoothness and second-moment constants $L_\Phi$ and $G_\Phi^2$ in the first two terms come from the gate ceiling $g_{n,I}\le 1$, since the per-layer gradient on the individual expert factors as $g_{n,I}\,\Phi_n^{(\ell)}$ in FLAME's architecture (Appendix Lemma~\ref{lem:jacobian}). 
\heng{The third term couples each client to the others:
its frozen experts are aggregated from other clients'
uploaded factors, and their inter-round changes enter
the bound through $2\mathcal{C}_S\Delta_T/(\eta K)$.
Under the stable consensus row-space condition,
the Grassmannian aggregate drifts no more than the
Euclidean mean (Lemma~\ref{lem:grass-drift}), tightening
this coupling term.} The gate floor $g_{\min}>0$, enforced by FLAME's enlarged top-$(M{+}1)$ router but not by FedALT's two-expert mixer, does not appear in \eqref{eq:thm-body}; it instead supplies the per-token Fisher mass on $(A_n,B_n)$ that the post-hoc Laplace stage of Section~\ref{sec:method-laplace} requires (Appendix Corollary~\ref{cor:fisher}).

\vspace{-0.5em}
\section{Experiments}
\label{sec:experiments}
\vspace{-0.5em}
We evaluate FLAME against personalized federated LoRA baselines and Bayesian baselines, and ablate the contribution of each FLAME component.

\vspace{-0.5em}
\subsection{Setup}
\label{sec:exp-setup}

We evaluate on four federated benchmarks: \emph{MCQA-12c}, a $12$-client FLAN classification setup over four tasks (3 IID clients each) on Llama-3.2-1B; and three $8$-client NLG generation setups on Llama-2-7B under increasing heterogeneity, namely \emph{FLAN} (IID replicas across four tasks), \emph{FLAN-XD} (cross-dataset within each task), and \emph{FLAN-LS} (disjoint label partitions within each task). All methods use LoRA rank $r=8$ at \texttt{q\_proj} and \texttt{v\_proj}, $T=10$ rounds with $T_w=1$ warmup, and the post-hoc KFAC Laplace runs once per client with $S=10$ posterior draws and adds no communication. We compare against FedAvg \citep{mcmahan2017communication}, FedDPA \citep{long2024dual}, FedALT \citep{fedalt}, FedLEASE \citep{fedlease}, and FLoRG \citep{florg2026} under matched training budget, and report accuracy / ROUGE-1, ECE, and NLL averaged across clients with mean $\pm$ standard deviation across seeds. \heng{For FLAME, classification accuracy is computed
from the MAP prediction, while ECE and NLL evaluate
the posterior-predictive distribution in
\eqref{eq:bma}.} Full setup details appear in Appendix~\ref{app:exp-datasets}.

\begin{table}[t]
\centering
\caption{Main results across four federated benchmarks. MCQA-12c is a $12$-client classification setup (Llama-3.2-1B); FLAN, FLAN-XD, and FLAN-LS are $8$-client NLG setups (Llama-2-7B) under IID, cross-dataset, and label-skew heterogeneity, respectively. The primary metric is accuracy on MCQA-12c and ROUGE-1 on the three FLAN benchmarks; ECE and NLL are reported on all four. All methods use LoRA rank $r=8$. \textbf{Bold} denotes the best value in each column.
\heng{Results are reported over three seeds.
For entries with standard deviations of $0.000$,
the three seeds differ by less than $0.001$,
yielding zero after rounding to three decimal places.}}
\label{tab:main}
\resizebox{\linewidth}{!}{%
\begin{tabular}{llc|ccc}
\toprule
 &  & MCQA-12c & \multicolumn{3}{c}{NLG-8c} \\
\cmidrule(lr){3-3} \cmidrule(lr){4-6}
Metric & Method & FLAN & FLAN & FLAN-XD & FLAN-LS \\
\midrule
\multirow{7}{*}{Acc / R-1 ($\uparrow$)}
& FedAvg                       & $73.58 \pm 0.55$ & $67.40 \pm 3.71$ & $69.15 \pm 6.80$ & $63.36 \pm 0.54$ \\
& FedDPA                       & $78.72 \pm 1.71$ & $70.25 \pm 3.87$ & $75.33 \pm 1.77$ & $70.34 \pm 3.25$ \\
& FedLEASE                     & $79.00 \pm 0.98$ & $73.17 \pm 3.37$ & $77.56 \pm 1.63$ & $74.69 \pm 1.04$ \\
& FedALT                       & $79.06 \pm 2.22$ & $73.03 \pm 4.66$ & $77.01 \pm 0.37$ & $73.76 \pm 2.13$ \\
& FLoRG                        & $72.61 \pm 0.51$ & $69.81 \pm 0.41$ & $72.51 \pm 0.28$ & $65.44 \pm 0.57$ \\
& \textbf{FLAME}               & $\mathbf{80.86 \pm 0.35}$ & $\mathbf{76.03 \pm 3.49}$ & $\mathbf{79.68 \pm 0.55}$ & $\mathbf{77.28 \pm 0.40}$ \\
\midrule
\multirow{7}{*}{ECE ($\downarrow$)}
& FedAvg                       & $0.212 \pm 0.008$ & $0.147 \pm 0.004$ & $0.167 \pm 0.001$ & $0.189 \pm 0.004$ \\
& FedDPA                       & $0.199 \pm 0.016$ & $0.195 \pm 0.011$ & $0.174 \pm 0.021$ & $0.199 \pm 0.003$ \\
& FedLEASE                     & $0.208 \pm 0.009$ & $0.176 \pm 0.001$ & $0.159 \pm 0.013$ & $0.205 \pm 0.001$ \\
& FedALT                       & $0.207 \pm 0.023$ & $0.181 \pm 0.011$ & $0.130 \pm 0.015$ & $0.184 \pm 0.002$ \\
& FLoRG                        & $0.212 \pm 0.004$ & $0.163 \pm 0.015$ & $0.161 \pm 0.012$ & $0.170 \pm 0.010$ \\
& \textbf{FLAME}               & $\mathbf{0.064 \pm 0.004}$ & $\mathbf{0.065 \pm 0.000}$ & $\mathbf{0.050 \pm 0.002}$ & $\mathbf{0.071 \pm 0.014}$ \\
\midrule
\multirow{7}{*}{NLL ($\downarrow$)}
& FedAvg                       & $1.28 \pm 0.08$ & $1.864 \pm 0.078$ & $1.747 \pm 0.056$ & $1.880 \pm 0.054$ \\
& FedDPA                       & $1.65 \pm 0.19$ & $1.828 \pm 0.160$ & $1.747 \pm 0.140$ & $1.941 \pm 0.300$ \\
& FedLEASE                     & $2.96 \pm 0.25$ & $2.216 \pm 0.130$ & $1.998 \pm 0.187$ & $2.265 \pm 0.100$ \\
& FedALT                       & $2.76 \pm 0.39$ & $2.087 \pm 0.280$ & $2.125 \pm 0.182$ & $2.196 \pm 0.360$ \\
& FLoRG                        & $1.294 \pm 0.016$ & $1.836 \pm 0.021$ & $1.793 \pm 0.015$ & $1.940 \pm 0.024$ \\
& \textbf{FLAME}               & $\mathbf{0.791 \pm 0.016}$ & $\mathbf{0.745 \pm 0.054}$ & $\mathbf{0.682 \pm 0.016}$ & $\mathbf{0.727 \pm 0.000}$ \\
\bottomrule
\end{tabular}%
}
\vspace{-0.5cm}
\end{table}

\subsection{Main results}
\label{sec:exp-main}

Table~\ref{tab:main} reports the full comparison across the four benchmarks. FLAME achieves the highest accuracy / ROUGE-1, the lowest ECE, and the lowest NLL on every benchmark. The accuracy lead is consistent with the hierarchical locality of Observation~\ref{obs:A} that existing personalized routers expose at a single level only. The calibration lead is substantially larger than the accuracy lead, in line with Observation~\ref{obs:calib} that the calibration gap is a property of point estimation rather than of any specific routing architecture. The ROUGE-1 lead persists across IID, cross-dataset, and label-skew heterogeneity regimes, indicating that the design choices motivated by the MCQA-12c parameter-space study transfer to generation tasks at a larger backbone scale. \heng{In our appendix, FLAME also retains the highest accuracy and
the lowest ECE and NLL when increasing the number
of clients from 12 to 16 under the same protocol
(Table~\ref{tab:mcqa16} in
Appendix~\ref{app:exp-scale}).} \heng{Across seven seeds on MCQA-12c, FLAME's accuracy
improvements over FedAvg, FedDPA, FedALT, and FedLEASE
are statistically significant under both one-sided
Wilcoxon signed-rank and paired $t$-tests
(Table~\ref{tab:significance} in
Appendix~\ref{app:exp-significance}).} \heng{On the newer Qwen3-1.7B-Base backbone, FLAME
again achieves the highest accuracy and the lowest ECE
among the evaluated methods
(Table~\ref{tab:qwen3} in Appendix~\ref{app:exp-backbone}).}

\vspace{-0.6em}\subsection{Ablations}
\label{sec:exp-ablation}

We isolate the contribution of each FLAME component on MCQA-12c (Table~\ref{tab:abl_component} in
Appendix~\ref{app:exp-component}): removing the post-hoc Laplace (\emph{w/o Lap.}) leaves accuracy unchanged but inflates ECE and NLL sharply, confirming Observation~\ref{obs:calib} that the calibration gap is closed only by the Bayesian model average; replacing it with MC Dropout (\emph{MCD}) does not recover FLAME's calibration; and further removing the Grassmannian aggregator (\emph{w/o Grass, w/o Lap.}) drops accuracy and inflates cross-seed variance, consistent with the gauge-invariance argument of Section~\ref{sec:method-grass}. Additional ablations corroborating these findings are reported in the appendix: hierarchy and aggregator ablations confirming Observations~\ref{obs:A}-\ref{obs:B} (Appendix~\ref{app:exp-arch}, Table~\ref{tab:abl_arch}), sanity controls ruling out random frozen experts and high-rank FedALT (Appendix~\ref{app:exp-control}, Table~\ref{tab:abl_control}), Laplace universality across baselines (Appendix~\ref{app:exp-lap-universal}, Table~\ref{tab:lap_universal}), and cluster recovery on FLAN-XD (Appendix~\ref{app:exp-cluster}). \heng{With the same post-hoc Laplace stage available
to every baseline, FLAME achieves the lowest ECE and
NLL, including against FLAME without the Grassmannian
aggregator (Table~\ref{tab:lap_universal}), demonstrating
the contribution of FLAME's architecture to calibration.}

\textbf{Additional experiments.}
Additional experiments in Appendix~\ref{app:exp-extra} include the
Local-only and additional federated LoRA comparisons
(Table~\ref{tab:local-only}), temperature-scaling controls
(Table~\ref{tab:temperature-scaling}), system-overhead measurements
(Tables~\ref{tab:communication-storage} and~\ref{tab:map-latency}),
and a 16-client scalability study (Table~\ref{tab:mcqa16}).
\heng{On Llama-2-7B, FLAME maintains MAP latency and peak GPU
memory close to those of FedLEASE, with posterior fitting
and sampling performed locally
(Tables~\ref{tab:map-latency-7b} and
\ref{tab:posterior-overhead-7b} in
Appendix~\ref{app:exp-overhead}).}

\section{\heng{Related Work}}
\label{sec:related}

\noindent\heng{\textbf{Federated LoRA.}
Parameter-efficient fine-tuning enables LLM adaptation
with a small trainable parameter footprint
\citep{ding2023parameter,fu2023effectiveness,hu2022lora}.
Recent studies extend LoRA to federated settings
\citep{zhang2023building,kuang2024federatedscope,
ye2024openfedllm,fan2023fate}.
FlexLoRA and FLoRA address heterogeneous adapter ranks
\citep{bai2024flexlora,wang2024flora}, while other works
study the aggregation error induced by separately
averaging the two LoRA factors
\citep{florg2026,sun2024improving}.
FLAME addresses the structural asymmetry of the two
factors under task heterogeneity by aligning $A$
on the Grassmann manifold while retaining Euclidean
aggregation on $B$.}

\smallskip
\noindent\heng{\textbf{Personalized federated LoRA.}
FedDPA blends global and local adapters at inference
\citep{long2024dual}.
FedALT introduces learned routing between an individual
expert and a frozen rest-of-world expert \citep{fedalt},
while FedLEASE constructs a clustered mixture of experts
\citep{fedlease}.
These methods expose a single level of cross-client
locality.
In contrast, FLAME jointly routes over individual,
intra-cluster, and cross-cluster experts, with the
trainable individual expert participating in every
forward pass.
This hierarchy and the selective aggregator follow
the measured locality structure and factor asymmetry
in Observations~\ref{obs:A} and~\ref{obs:B}.}

\smallskip
\noindent\heng{\textbf{Bayesian LoRA and uncertainty quantification.}
Bayesian LoRA methods investigate predictive uncertainty
in centralized LLM adaptation
\citep{wang2024blob,yang2023bayesian,wang2023lora,
tian2024variational}, while federated uncertainty
quantification has primarily focused on smaller models
\citep{achituve2021personalized,kassab2022federated}.
FLAME integrates a local post-hoc Kronecker-factored
Laplace posterior into personalized federated LoRA,
leaving training and per-round communication unchanged.
The guaranteed participation of the individual expert
ensures that its posterior samples enter the prediction
at every token.
Figure~\ref{fig:method_compare} compares FLAME with
representative federated LoRA methods, and
Appendix~\ref{app:related} provides more.}

\section{Conclusion}
\label{sec:conclusion}

We introduced FLAME, a personalized federated LoRA framework whose three components, a clustered mixture of experts with a guaranteed individual path, a selective Grassmannian aggregator on the input-side factor, and a post-hoc Kronecker-factored Laplace posterior, are each motivated by a measured property of the trained parameter space. A non-asymptotic convergence analysis attributes a distinct term of the stationary-point bound to each component. Across a 12-client MCQA benchmark and three FLAN NLG benchmarks under varying heterogeneity, FLAME improves accuracy and calibration over strong personalized federated LoRA baselines without adding per-round communication. The post-hoc Laplace stage is currently restricted to a Kronecker-factored Gaussian over the individual expert, and the Grassmannian aggregator assumes a shared rank across clients. Extending the posterior family and integrating the aggregator with heterogeneous-rank federated LoRA are natural directions for future work.

\clearpage

\section*{Reproducibility Statement}
The complete algorithm is provided in Algorithm~\ref{alg:flame}.
Experimental settings and additional results are provided in
Appendix~\ref{app:exp-extra}, and the complete convergence analysis
is provided in Appendix~\ref{sec:convergence}.

\section*{Ethics Statement}
This work evaluates federated fine-tuning methods on existing
language-model benchmarks.

\section*{AI Use Statement}
Generative AI tools were used to assist with language editing\haotian{, making the flowcharts} and
manuscript organization.

\clearpage
\begin{center}
    \textbf{\Large Supplementary File}
\end{center}
\appendix
\section{Related Work}
\label{app:related}

\paragraph{Parameter-Efficient Fine-Tuning and LoRA.}
As LLMs grow in scale, full-parameter fine-tuning becomes prohibitive in compute and memory \citep{hadi2023survey}. Parameter-efficient fine-tuning methods \citep{ding2023parameter, fu2023effectiveness} reduce the trainable parameter footprint by introducing a small set of additional parameters while keeping the pretrained backbone frozen. LoRA \citep{hu2022lora, li2026low} approximates the fine-tuning update by a low-rank matrix $BA$, with $A$ Kaiming-initialized and $B$ zero-initialized so that the residual is exactly zero at the start of training. Subsequent work improves LoRA along multiple axes, including asymmetric architectures \citep{tian2024hydra}, mixture-of-experts variants \citep{liu2023moelora, dou2024loramoe}, and rank-adaptive methods \citep{zhang2023adalora}. These advances are developed primarily for centralized training, and the question of how to deploy LoRA across clients with heterogeneous data distributions remains open.

\paragraph{Federated Learning and Federated Fine-Tuning.}
Federated Averaging (FedAvg) \citep{mcmahan2017communication} is the foundational FL algorithm and aggregates locally trained models on the server before redistributing to clients. Personalized federated learning \citep{tan2022towards, collins2021exploiting, fallah2020personalized} addresses statistical heterogeneity by allowing clients to retain client-specific parameters, but most existing methods target small models and label-distribution skew rather than LLMs and task heterogeneity. The intersection of FL and PEFT has produced a growing federated fine-tuning literature \citep{zhang2023building, kuang2024federatedscope, ye2024openfedllm, fan2023fate}, with FedIT \citep{zhang2023building} pioneering federated LoRA, FlexLoRA \citep{bai2024flexlora}, and FLoRA \citep{wang2024flora} addressing heterogeneous client ranks, and a separate line analyzing the bilinear aggregation error of separately averaging $A$ and $B$ \citep{florg2026, sun2024improving}. These methods either share a single global LoRA across all clients or address aggregation under the assumption of label-shift heterogeneity, and do not target the task-heterogeneous regime that is the natural deployment scenario for LLMs.

\paragraph{Personalized Federated LoRA and Uncertainty Quantification.}
The most relevant prior works personalize federated LoRA under task heterogeneity. FedDPA \citep{long2024dual} maintains a global LoRA and a local LoRA per client and blends them at inference with a fixed mixing weight. FedALT \citep{fedalt} replaces the fixed weight with a learned softmax router and treats the average over all other clients as a frozen RoW LoRA. FedLEASE \citep{fedlease} replaces the single RoW by a clustered mixture of experts whose number is selected by silhouette score on the cosine similarity of LoRA $B$-matrices. These designs each operate at a single level of cross-client locality, either binary or per-cluster, and apply Euclidean averaging to both LoRA factors regardless of how each factor carries signal across clients. None of them addresses the calibration of the resulting per-client adapters. Centralized Bayesian LoRA methods \citep{wang2024blob, yang2023bayesian, wang2023lora, tian2024variational} provide calibrated predictive distributions in the centralized regime, but each modifies the training loop or adds per-step memory cost that does not scale across federated clients. Federated UQ work \citep{achituve2021personalized, kassab2022federated} targets sub-LLM scale models and does not adapt to the LoRA regime. Personalized federated LLMs therefore inherit no usable uncertainty quantification recipe, which is the gap our local post-hoc Laplace closes. FLAME extends this line of work along three axes that prior personalized federated LoRA designs leave unaddressed. Where FedALT and FedLEASE expose only a binary or per-cluster split, FLAME routes among the trainable individual adapters, an intra-cluster aggregate, and cross-cluster aggregates jointly, exposing the hierarchical locality the trained parameter space exhibits. Where prior designs apply Euclidean averaging to both LoRA factors, FLAME aligns $A$ on the Grassmann manifold to remove the basis ambiguity while keeping $B$ in Euclidean space. Where prior designs return point-estimate adapters, FLAME equips each client with a post-hoc Kronecker-factored Laplace posterior whose computation is local and adds no communication.

\begin{figure*}[t]
\centering
\includegraphics[width=0.98\textwidth]{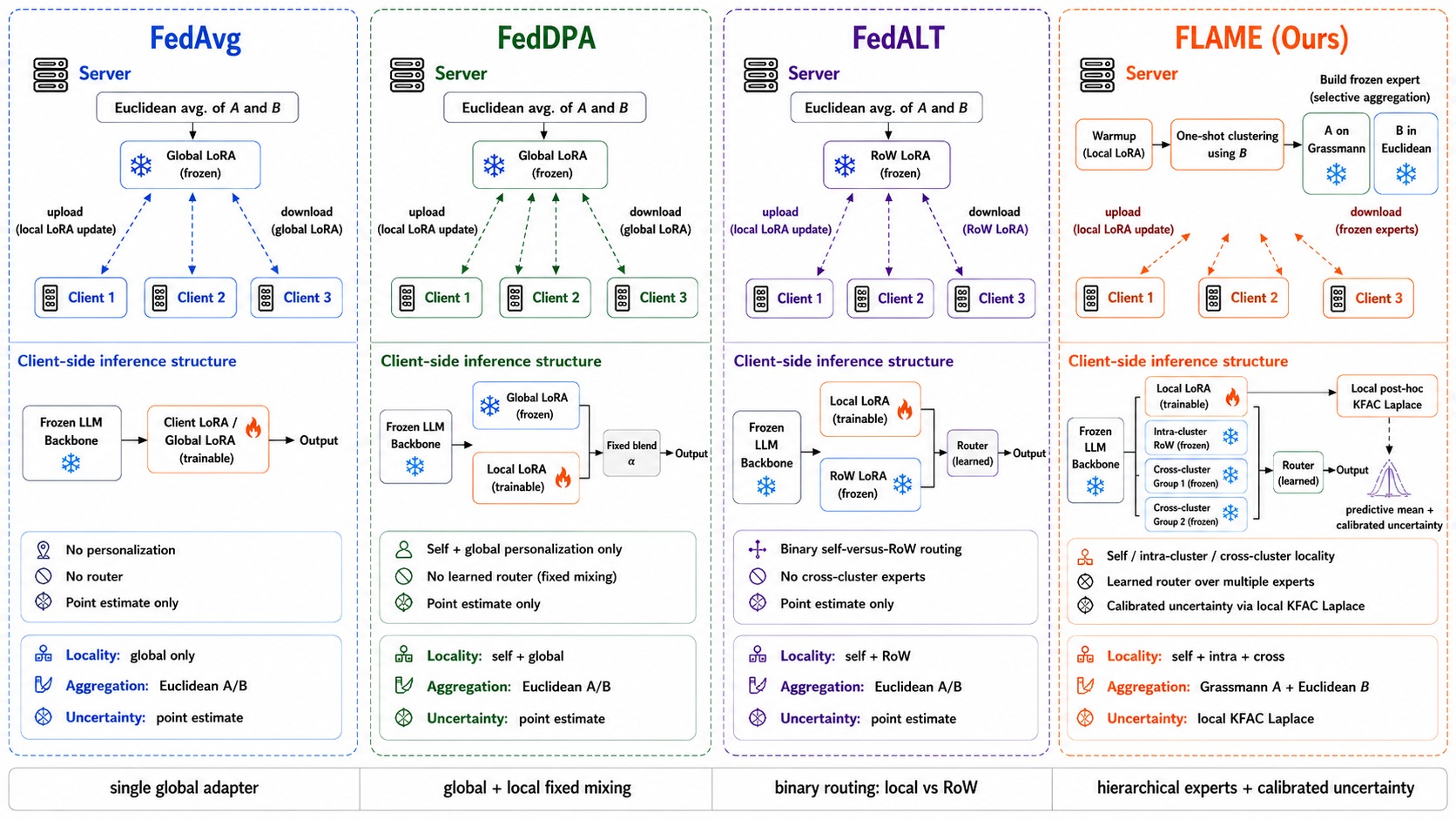}
\caption{
Comparison of personalized federated LoRA methods. 
FedAvg uses a single global adapter shared across clients. 
FedDPA combines a global adapter with a local adapter through fixed mixing. 
FedALT routes between a client-specific local adapter and a single Row adapter. 
FLAME jointly exploits self, intra-cluster, and cross-cluster locality, aggregates the input-side factor $A$ on the Grassmann manifold while keeping Euclidean aggregation on the output-side factor $B$, and adds local post-hoc KFAC Laplace to obtain calibrated predictive uncertainty.
}
\label{fig:method_compare}
\end{figure*}

\section{Additional Experimental Details and Results}
\label{app:exp-extra}

\subsection{Full Setup Details}
\label{app:exp-datasets}

\paragraph{Datasets and clients.} \emph{MCQA-12c} is a $12$-client FLAN classification setup with four tasks (3 IID clients each): SNLI (3-class NLI), HellaSwag (4-class commonsense), BoolQ (2-class reading comprehension), and Sentiment140 (2-class sentiment). The remaining three benchmarks are $8$-client FLAN generation setups (NLG-8c) with four tasks (2 clients each) under increasing heterogeneity. \emph{FLAN} assigns two IID replicas per task across four tasks: Sentiment140 (sentiment), WebNLG (struct-to-text), FixPunct (text editing), and HellaSwag (commonsense). \emph{FLAN-XD} replaces each IID replica with a distinct source dataset from the same task category, yielding eight datasets: Sentiment140/SST2, WebNLG/DART, FixPunct/WordSegment, and HellaSwag/PIQA. \emph{FLAN-LS} uses the same four datasets as FLAN but splits each task into two clients with disjoint label partitions (e.g., positive-only vs negative-only for Sentiment140). Each client holds 600 training and 100 test samples; data are disjoint across clients in all settings.

\paragraph{Backbone and adapter.} MCQA-12c uses Llama-3.2-1B; the three NLG-8c benchmarks use Llama-2-7B. Both backbones are loaded in 8-bit quantization via \texttt{bitsandbytes} (LLM.int8), with LoRA weights computed in \texttt{float16}. LoRA rank $r=8$ is applied to the \texttt{q\_proj} and \texttt{v\_proj} layers of every transformer block, with $\alpha=16$ and dropout $0.05$.

\paragraph{Federated protocol.} The local warmup runs $T_w=1$ round of local-only training without aggregation. The server then clusters clients by agglomerative clustering on the entry-level $B$-similarity distance \eqref{eq:cluster_dist}, selecting $M$ by silhouette score (recovering $M=4$ on all benchmarks, matching the underlying number of tasks). Clustered MoE training runs $T-T_w=9$ additional rounds (total $T=10$ communication rounds), with $E=10$ local epochs per round on MCQA-12c and $E=1$ on NLG-8c. The post-hoc KFAC Laplace stage runs once per client on its training data with $S=10$ posterior draws; no additional communication is required.

\paragraph{Optimization and random seeds.}
We use AdamW with learning rate $10^{-4}$, weight decay $0.05$, and
batch size $4$. All main experiments and ablations are reported over
three random seeds.

\paragraph{Baselines.} We compare against FedAvg \citep{mcmahan2017communication}, FedDPA \citep{long2024dual}, FedALT \citep{fedalt}, FedLEASE \citep{fedlease}, and FLoRG \citep{florg2026}, all under the same LoRA rank, training rounds, and per-round local epochs.

\paragraph{Metrics.} On MCQA-12c, we report classification accuracy, expected calibration error (ECE), and negative log-likelihood (NLL). \heng{FLAME's classification accuracy is that of
the MAP prediction, while ECE and NLL are computed
from the posterior-predictive distribution in
\eqref{eq:bma}.} On the three NLG-8c benchmarks, we report ROUGE-1 and per-token ECE and NLL. Each metric is computed per client and averaged across clients. Performances are reported as mean $\pm$ standard deviation across multiple random seeds.

\subsection{Component Ablation}
\label{app:exp-component}

\begin{table}[t]
\centering
\caption{Component ablation on MCQA-12c. We compare the full FLAME against variants that swap or disable components: \emph{w/o Grass, w/o Lap.} uses Euclidean aggregation and MAP prediction; \emph{w/o Lap.} adds Grassmannian $A$ but keeps MAP; \emph{MCD} replaces the KFAC Laplace with MC Dropout ($K{=}10$). All variants use LoRA rank $r{=}8$. \textbf{Bold} denotes the best in each column. \heng{Classification accuracy for FLAME is evaluated
at the MAP prediction and is therefore unchanged
by the post-hoc Laplace stage.}}
\label{tab:abl_component}
\begin{tabular}{lccc}
\toprule
Variant & Acc (\%) $\uparrow$ & ECE $\downarrow$ & NLL $\downarrow$ \\
\midrule
FLAME (w/o Grass, w/o Lap.) & $79.83 \pm 1.41$ & $0.212 \pm 0.007$ & $2.500 \pm 0.291$ \\
FLAME (w/o Lap.)            & $80.86 \pm 0.35$ & $0.204 \pm 0.013$ & $2.359 \pm 0.217$ \\
FLAME (MCD)                 & $77.62 \pm 1.88$ & $0.202 \pm 0.015$ & $2.166 \pm 0.151$ \\
\textbf{FLAME}              & $\mathbf{80.86 \pm 0.35}$ & $\mathbf{0.064 \pm 0.004}$ & $\mathbf{0.791 \pm 0.016}$ \\
\bottomrule
\end{tabular}
\vspace{-0.6cm}
\end{table}

\subsection{Dataset Specifications}
\label{app:exp-datasets}

We evaluate on four federated benchmarks spanning classification and generation. \emph{MCQA-12c} is a $12$-client FLAN classification setup with four tasks (3 IID clients each): SNLI (3-class NLI), HellaSwag (4-class commonsense), BoolQ (2-class reading comprehension), and Sentiment140 (2-class sentiment). The remaining three benchmarks are $8$-client FLAN generation setups (NLG-8c) with four tasks (2 clients each) under increasing heterogeneity. \emph{FLAN} assigns two IID replicas per task across four tasks: Sentiment140 (sentiment), WebNLG (struct-to-text), FixPunct (text editing), and HellaSwag (commonsense). \emph{FLAN-XD} replaces each IID replica with a distinct source dataset from the same task category, yielding eight datasets: Sentiment140/SST2, WebNLG/DART, FixPunct/WordSegment, and HellaSwag/PIQA. \emph{FLAN-LS} uses the same four datasets as FLAN but splits each task into two clients with disjoint label partitions (e.g., positive-only vs negative-only for Sentiment140). Each client holds 600 training and 100 test samples; data are disjoint across clients in all settings.

\subsection{Local-Only and Additional Federated LoRA Baselines}
\label{app:exp-local}

We extend the MCQA-12c comparison with a matched Local-only reference
and two additional federated LoRA baselines, FFA-LoRA and LoRM.
For Local-only, each client trains only its individual LoRA on its own
600 samples. Communication, clustering, routing, and all shared experts
are removed, while the initialization, optimizer, round boundaries,
local epochs, batch size, and total local optimization budget are
matched to the federated methods.

\begin{table}[h]
\centering
\small
\caption{Local-only and additional federated LoRA baselines on MCQA-12c.}
\label{tab:local-only}
\begin{tabular}{lccc}
\toprule
Method & Acc. (\%) $\uparrow$ & ECE $\downarrow$ & NLL $\downarrow$ \\
\midrule
FedAvg
& $73.58 \pm 0.55$
& $0.212 \pm 0.008$
& $1.28 \pm 0.08$ \\
FLoRG
& $72.61 \pm 0.51$
& $0.212 \pm 0.004$
& $1.29 \pm 0.02$ \\
FFA-LoRA
& $61.22 \pm 0.56$
& $0.216 \pm 0.011$
& $1.04 \pm 0.01$ \\
LoRM
& $69.69 \pm 0.87$
& $0.242 \pm 0.003$
& $1.30 \pm 0.08$ \\
Local-only
& $78.45 \pm 1.59$
& $0.203 \pm 0.018$
& $1.62 \pm 0.24$ \\
FedDPA
& $78.72 \pm 1.71$
& $0.199 \pm 0.016$
& $1.65 \pm 0.19$ \\
FedLEASE
& $79.00 \pm 0.98$
& $0.208 \pm 0.009$
& $2.96 \pm 0.25$ \\
FedALT
& $79.06 \pm 2.22$
& $0.207 \pm 0.023$
& $2.76 \pm 0.39$ \\
FLAME
& $80.86 \pm 0.35$
& $0.064 \pm 0.004$
& $0.791 \pm 0.016$ \\
\bottomrule
\end{tabular}
\end{table}

The single-global-adapter methods, including FFA-LoRA and LoRM,
underperform Local-only by $4.87$--$17.23$ accuracy points.
FedDPA, FedLEASE, and FedALT exceed Local-only accuracy by only
$0.27$, $0.55$, and $0.61$ points, respectively, which are small
relative to their across-seed variation. None improves both calibration
metrics over Local-only. In contrast, FLAME improves accuracy by
$2.41$ points while reducing ECE from $0.203$ to $0.064$ and NLL
from $1.62$ to $0.791$. The comparison shows that federation is useful
in this setting when the method preserves client-specific adaptation
and selectively shares compatible cross-client information.

\subsection{Temperature Scaling}
\label{app:exp-ts}

We evaluate temperature scaling (TS) on MCQA-12c under two calibration
protocols.

\paragraph{Protocol A: matched label access.}
The temperature $T$ is fitted on the same local training labels
available to FLAME's evidence maximization, and every method is
evaluated on the full test split. Values are means over the three
training seeds.

\begin{table}[h]
\centering
\small
\caption{Temperature scaling on MCQA-12c under matched label access.}
\label{tab:temperature-scaling}
\begin{tabular}{lcccc}
\toprule
Method
& ECE per-cl.\ $T$
& ECE global $T$
& NLL per-cl.\ $T$
& NLL global $T$ \\
\midrule
FedAvg
& 0.098 & 0.116 & 0.612 & 0.636 \\
FLoRG
& 0.098 & 0.107 & 0.629 & 0.652 \\
FFA-LoRA
& 0.087 & 0.123 & 0.748 & 0.779 \\
LoRM
& 0.090 & 0.149 & 0.686 & 0.742 \\
FedDPA
& 0.210 & 0.209 & 14.50 & 10.37 \\
FedLEASE
& 0.203 & 0.203 & 3.88 & 4.57 \\
FedALT
& 0.209 & 0.208 & 2.63 & 2.13 \\
Local-only
& 0.214 & 0.215 & 5.52 & 5.92 \\
\bottomrule
\end{tabular}
\end{table}

Under this matched comparison, the best temperature-scaled baseline
reaches ECE $0.087$, compared with FLAME's $0.064$.

\paragraph{Protocol B: held-out sensitivity check.}
The temperature $T$ is fitted on half of each client's test split and
the temperature-scaled baselines are evaluated on the disjoint half.
FLAME uses neither half for calibration. The deterministic 50/50
client-level split uses a fixed RNG seed and is identical across
methods and training seeds.

The best per-client temperature-scaled ECE/NLL results are
$0.086/0.562$ for FedALT, $0.087/0.537$ for FedLEASE,
$0.104/0.536$ for FedDPA, $0.105/0.534$ for Local-only,
$0.099/0.598$ for FLoRG, and $0.108/0.584$ for FedAvg.
The best global-$T$ ECE is $0.123$.

Because Protocol B changes both calibration-label access and the
evaluated subset, these half-test results are reported as a sensitivity
check rather than compared directly with FLAME's full-test result.
FLAME's Laplace stage uses only the training data already available to
each client and requires no separate calibration set. Under the matched
training-label protocol, FLAME reaches ECE $0.064$, compared with
$0.087$ for the best train-fitted temperature-scaled baseline.

\subsection{System Overhead}
\label{app:exp-overhead}

We profile communication, storage, computation, memory, and inference
latency on MCQA-12c using Llama-3.2-1B on an RTX 6000 Ada.

\paragraph{Communication and storage.}

\begin{table}[h]
\centering
\small
\caption{Communication and persistent model state on MCQA-12c.}
\label{tab:communication-storage}
\begin{tabular}{lccc}
\toprule
Method & Uplink & Downlink & Persistent model state \\
\midrule
Local-only
& 0 & 0 & 1.62 MB \\
FFA-LoRA
& 0.62 MB & 0.62 MB & 1.62 MB \\
LoRM
& 0.93 MB & 1.62 MB & 1.62 MB \\
FedAvg / FLoRG
& 1.62 MB & 1.62 MB & 1.62 MB \\
FedALT
& 1.62 MB & 1.62 MB & 1.62 MB \\
FedDPA
& 1.62 MB & 1.62 MB & 3.25 MB \\
FedLEASE
& 1.62 MB & 6.50 MB & 9.00 MB \\
FLAME
& 1.62 MB & 6.50 MB & 9.25 MB \\
\bottomrule
\end{tabular}
\end{table}

These figures use rank-$8$ fp16 LoRA states. FLAME uploads one
$(A_n,B_n)$ pair per client, giving the same $1.62$ MB uplink as
FedAvg, FedALT, and FedDPA. Its $6.50$ MB downlink and $9.25$ MB
persistent model state arise from the routed expert set and are close
to FedLEASE's $6.50$ MB and $9.00$ MB. The post-hoc Laplace stage is
entirely local and adds no communication. At fixed cluster count $M$,
the per-client expert footprint is $\mathcal{O}(M)$ rather than
$\mathcal{O}(N)$.

\paragraph{Profiling and inference latency.}

\begin{table}[h]
\centering
\small
\caption{Single-forward MAP inference latency on MCQA-12c.}
\label{tab:map-latency}
\begin{tabular}{llc}
\toprule
Method & Structure & ms/sample \\
\midrule
LoRM
& Single adapter & 9.85 \\
FedAvg
& Single adapter & 9.98 \\
FLoRG / Local-only
& Single adapter & 10.04 \\
FedDPA
& Single adapter & 10.06 \\
FedALT
& RoW-routed & 10.33 \\
FFA-LoRA
& Single adapter & 10.38 \\
FedLEASE
& MoE & 12.09 \\
FLAME
& MoE & 12.36 \\
\bottomrule
\end{tabular}
\end{table}

FLAME's MAP path is $24\%$ slower than FedAvg and close to FedLEASE.
The one-time post-hoc KFAC fitting stage takes $233.7$ s per client and
consists of one forward-backward profiling sweep without parameter
updates. Its measured temporary fp32 memory peak is $784$ MB.

In the offline evaluation, the local posterior state is retained through
posterior prediction and released afterward. Continuous calibrated
serving would instead retain the local factors or eigendecompositions,
or cache posterior samples; this state remains local but is additional
to the $9.25$ MB MAP model state reported above.

Posterior-predictive latency grows approximately linearly with the
sample count $S$. Latencies are $13.1$, $26.3$, $65.7$, and
$131.9$ ms per prediction for $S=1,2,5,$ and $10$, respectively,
averaged over the 12 clients. Latency is profiled using seed 36.
The main experiments use $S=10$, for which FLAME reports ECE
$0.064 \pm 0.004$ and NLL $0.791 \pm 0.016$ over three seeds.

\begin{table}[h]
\centering
\addtocounter{table}{-1}%
\renewcommand{\thetable}{6 (Llama-2-7B MAP)}%
\renewcommand{\theHtable}{6map7b}%
\caption{
\heng{MAP inference latency and peak GPU memory
on Llama-2-7B.}
}
\label{tab:map-latency-7b}

\begin{tabular*}{\linewidth}{
@{\extracolsep{\fill}}lcc@{}}
\toprule
Method & Latency (ms/sequence) & Peak GPU memory (MB) \\
\midrule
Single adapter & 22.86 & 8,108 \\
FedALT         & 24.47 & 8,108 \\
FedLEASE       & 27.08 & 8,127 \\
FLAME          & 28.35 & 8,182 \\
\bottomrule
\end{tabular*}
\end{table}

\heng{Table~\ref{tab:map-latency-7b} extends the MAP
inference comparison to Llama-2-7B.
FLAME maintains latency and peak GPU memory close to
those of FedLEASE, consistent with the results on
Llama-3.2-1B.}

\begin{table}[h]
\centering
\addtocounter{table}{-1}%
\renewcommand{\thetable}{6 (Llama-2-7B posterior)}%
\renewcommand{\theHtable}{6post7b}%
\caption{
\heng{Per-client posterior fitting and sampling
overhead on Llama-2-7B.}
}
\label{tab:posterior-overhead-7b}

\begin{tabular*}{\linewidth}{
@{\extracolsep{\fill}}lc@{}}
\toprule
Metric & Cost \\
\midrule
One-time posterior fit & 997 s \\
Drawing one posterior sample & 12.85 ms \\
Cached posterior samples for $S=10$ (fp16) & 80 MB \\
\bottomrule
\end{tabular*}
\end{table}

\heng{Table~\ref{tab:posterior-overhead-7b} reports
the posterior fitting and sampling costs on Llama-2-7B.
The posterior is fitted once per client through a
single forward-backward sweep without parameter updates
or additional communication.
For serving, caching $S$ posterior samples adds $S$
adapter-sized copies per client to the MAP model state.} \heng{At this cost, FLAME achieves the lowest ECE and
NLL even when all baselines use the same Laplace stage
(Table~\ref{tab:lap_universal}).}

\subsection{Client-Count Scalability}
\label{app:exp-scale}

We extend MCQA from 12 to 16 clients while keeping Llama-3.2-1B,
600 local training samples per client, and the training protocol
unchanged. Results are means and standard deviations over three seeds.

\begin{table}[h]
\centering
\small
\caption{MCQA results with 16 clients under the same backbone,
per-client data budget, and training protocol as MCQA-12c.}
\label{tab:mcqa16}
\begin{tabular}{lccc}
\toprule
Method & Acc. (\%) $\uparrow$ & ECE $\downarrow$ & NLL $\downarrow$ \\
\midrule
FedAvg
& $73.13 \pm 1.24$
& $0.210 \pm 0.013$
& $1.236 \pm 0.078$ \\
FedALT
& $78.75 \pm 0.84$
& $0.210 \pm 0.008$
& $3.179 \pm 0.126$ \\
FedLEASE
& $79.67 \pm 0.62$
& $0.201 \pm 0.006$
& $3.206 \pm 0.158$ \\
FLAME
& $80.58 \pm 0.61$
& $0.063 \pm 0.003$
& $0.761 \pm 0.026$ \\
\bottomrule
\end{tabular}
\end{table}

FLAME retains the highest accuracy and the lowest ECE and NLL when
increasing the client count from 12 to 16. Calibration remains stable
relative to MCQA-12c, where FLAME reports ECE $0.064 \pm 0.004$ and
NLL $0.791 \pm 0.016$.

\subsection{\heng{Statistical Significance}}
\label{app:exp-significance}

\heng{We extend the three seeds of Table~\ref{tab:main}
to seven seeds on MCQA-12c and evaluate FLAME's accuracy
against FedAvg, FedDPA, FedALT, and FedLEASE using two
one-sided paired tests.
The Wilcoxon signed-rank test uses 84 client-seed pairs,
while the paired $t$-test uses seven seeds.
Table~\ref{tab:significance} shows that FLAME's accuracy
improvements over all four baselines are statistically
significant under both tests.}

\begin{table}[h]
\centering
\addtocounter{table}{-1}%
\renewcommand{\thetable}{1 (Significance)}%
\renewcommand{\theHtable}{1sig}%
\caption{\heng{Paired statistical comparisons of accuracy
on MCQA-12c over seven seeds. Both tests are one-sided,
in the direction of FLAME improving on the baseline.}}
\label{tab:significance}

\begin{tabular*}{\linewidth}{
@{\extracolsep{\fill}}lcc@{}}
\toprule
FLAME vs.
& \shortstack{Wilcoxon signed-rank $p$\\
(84 client-seed pairs)}
& \shortstack{Paired $t$-test $p$\\(7 seeds)} \\
\midrule
FedAvg
& $2.8\times10^{-10}$
& $6.4\times10^{-5}$ \\
FedDPA
& $0.0047$
& $0.027$ \\
FedALT
& $0.011$
& $0.010$ \\
FedLEASE
& $0.028$
& $0.041$ \\
\bottomrule
\end{tabular*}
\end{table}

\subsection{\heng{Additional Backbone}}
\label{app:exp-backbone}

\heng{We further evaluate FLAME on MCQA-12c using
Qwen3-1.7B-Base under the same experimental protocol,
with results reported over three seeds.
Table~\ref{tab:qwen3} reports the comparison with
FedAvg, FedALT, and FedLEASE.
FLAME again achieves the highest accuracy and
the lowest ECE on the newer backbone.}

\begin{table}[h]
\centering
\addtocounter{table}{-1}%
\renewcommand{\thetable}{1 (Qwen3)}%
\renewcommand{\theHtable}{1qwen}%
\caption{\heng{Results on MCQA-12c with Qwen3-1.7B-Base
for FLAME and selected baselines, reported as mean
$\pm$ standard deviation over three seeds.
\textbf{Bold} denotes the best value in each column.}}
\label{tab:qwen3}

\begin{tabular*}{\linewidth}{
@{\extracolsep{\fill}}lccc@{}}
\toprule
Method
& Acc. (\%) $\uparrow$
& ECE $\downarrow$
& NLL $\downarrow$ \\
\midrule
FedAvg
& $85.83 \pm 2.05$
& $0.118 \pm 0.017$
& $\mathbf{0.69 \pm 0.08}$ \\
FedALT
& $86.33 \pm 0.80$
& $0.135 \pm 0.007$
& $1.77 \pm 0.10$ \\
FedLEASE
& $86.75 \pm 0.72$
& $0.133 \pm 0.006$
& $1.75 \pm 0.16$ \\
\textbf{FLAME}
& $\mathbf{87.22 \pm 1.28}$
& $\mathbf{0.086 \pm 0.004}$
& $0.87 \pm 0.11$ \\
\bottomrule
\end{tabular*}
\end{table}

\subsection{Architecture Ablations}
\label{app:exp-arch}

Table~\ref{tab:abl_arch} drills into each architectural choice on MCQA-12c. The hierarchy ablation in Table~\ref{tab:abl_arch}(a) removes one expert level at a time. Every removal degrades accuracy and inflates cross-seed variance, most sharply when the cross-cluster experts are removed; all three levels contribute, and none is redundant, in line with the three-granularity locality of Observation~\ref{obs:A}. The aggregator ablation in Table~\ref{tab:abl_arch}(b) crosses Grassmannian and Euclidean aggregation on $A$ and $B$ independently. Grassmannian on $A$ paired with Euclidean on $B$ outperforms every other combination, while the symmetric Grass-Grass and Eucl-Eucl variants both lag, confirming the asymmetry of Observation~\ref{obs:B}: $A$-subspaces carry shared structure that benefits from gauge-invariant alignment, whereas Grassmannian projection on $B$ disrupts the entry-level task identity it encodes.

\begin{table}[h]
\centering

\caption{Architecture ablations on MCQA-12c.
\textbf{(a) C1: hierarchy ablation} removes one
expert level at a time from FLAME, retaining the
Grassmannian $A$ aggregator.
\textbf{(b) C2: aggregator} crosses Grassmannian
(Grass) and Euclidean (Eucl) aggregation on $A$
and $B$ independently.
All variants use LoRA rank $r{=}8$.
\textbf{Bold} denotes the best in each panel.}
\label{tab:abl_arch}
\begin{subtable}[t]{0.49\linewidth}
\centering
\caption{C1: hierarchy ablation}
\label{tab:abl_arch_c1}
\begin{tabular}{lc}
\toprule
Variant & Acc (\%) $\uparrow$ \\
\midrule
\textbf{FLAME (all three levels)} & $\mathbf{80.86 \pm 0.35}$ \\
w/o cross-cluster experts & $79.47 \pm 1.71$ \\
w/o intra-cluster RoW & $79.00 \pm 0.76$ \\
w/o individual expert & $79.06 \pm 0.43$ \\
\bottomrule
\end{tabular}
\end{subtable}%
\hfill
\begin{subtable}[t]{0.49\linewidth}
\centering
\caption{C2: aggregator}
\label{tab:abl_arch_c2}
\begin{tabular}{llc}
\toprule
$A$ agg. & $B$ agg. & Acc (\%) $\uparrow$ \\
\midrule
\textbf{Grass} & \textbf{Eucl} & $\mathbf{80.86 \pm 0.35}$ \\
Grass & Grass & $79.64 \pm 0.51$ \\
Eucl & Grass & $79.14 \pm 1.21$ \\
Eucl & Eucl & $79.83 \pm 1.41$ \\
\bottomrule
\end{tabular}
\end{subtable}
\end{table}

\subsection{Sanity Controls}
\label{app:exp-control}

Two sanity controls in Table~\ref{tab:abl_control} address natural alternative explanations for FLAME's gain. The \emph{Random experts} variant replaces the content of FLAME's frozen experts with norm-matched random weights while keeping the routing architecture identical, and underperforms FLAME by a wide margin, isolating the gain to the content of the cluster-aware aggregates rather than the routing architecture alone. The \emph{FedALT ($r{=}40$)} variant raises FedALT's LoRA rank by a factor of five over its default; this lifts its accuracy to FLAME's level but does not close the calibration gap, so the gap cannot be closed by adding adapter parameters to a point-estimate routing scheme.

\begin{table}[h]
\centering
\caption{Sanity controls on MCQA-12c. \emph{Random experts} replaces the content of FLAME's frozen experts with norm-matched random weights while keeping the routing architecture identical to FLAME, isolating the value of the cluster-aware aggregates from the routing architecture. \emph{FedALT ($r{=}40$)} is a high-capacity baseline that raises FedALT's LoRA rank by a factor of five. \textbf{FLAME} is reported as the reference point.}
\label{tab:abl_control}
\begin{tabular}{lccc}
\toprule
Variant & Acc (\%) $\uparrow$ & ECE $\downarrow$ & NLL $\downarrow$ \\
\midrule
Random experts          & $78.46 \pm 0.65$ & $0.215 \pm 0.008$ & $3.14 \pm 0.11$ \\
FedALT ($r{=}40$)       & $80.25 \pm 0.82$ & $0.197 \pm 0.006$ & $2.58 \pm 0.11$ \\
\midrule
\textbf{FLAME}          & $\mathbf{80.86 \pm 0.35}$ & $\mathbf{0.064 \pm 0.004}$ & $\mathbf{0.791 \pm 0.016}$ \\
\bottomrule
\end{tabular}
\end{table}

\subsection{Universality of Post-hoc Laplace}
\label{app:exp-lap-universal}

A natural follow-up to the component ablation in Section~\ref{sec:exp-ablation} is whether the Bayesian gain transfers to any personalized federated LoRA design, or whether it is bound to FLAME's specific architecture.
\heng{Table~\ref{tab:lap_universal} reports per-client
best(MAP, Laplace) ECE and NLL after equipping every
baseline with the same post-hoc Laplace stage on
MCQA-12c. For each method and client, the better of
MAP and Laplace is selected on the evaluation labels,
using the same selection rule for FLAME and every
baseline. Under this protocol, FLAME achieves the
lowest ECE and NLL.} Among the baselines, FedALT is the closest architectural match to FLAME, sharing the trainable-individual-plus-frozen design but operating at a single locality scale; the persistent gap of FLAME over FedALT post-Laplace confirms that the calibration gain comes from the conjunction of FLAME's three-granularity routing, the selective Grassmannian aggregator, and the post-hoc Laplace, and is not reproducible by appending Laplace to a single-locality-scale individual-plus-frozen routing architecture alone.

\begin{table}[h]
\centering

\caption{\heng{Universality of post-hoc Laplace on
MCQA-12c. Each row reports per-client best
(MAP, Laplace) ECE and NLL under the same selection
rule. \textbf{FLAME} achieves the lowest ECE and NLL.}}
\label{tab:lap_universal}
\begin{tabular}{lcc}
\toprule
Method & Best ECE $\downarrow$ & Best NLL $\downarrow$ \\
\midrule
FedAvg & $0.079 \pm 0.006$ & $0.821 \pm 0.005$ \\
FedDPA & $0.086 \pm 0.007$ & $0.823 \pm 0.005$ \\
FedALT & $0.081 \pm 0.004$ & $0.821 \pm 0.002$ \\
FedLEASE & $0.086 \pm 0.013$ & $0.817 \pm 0.005$ \\
FLAME (w/o Grass) & $0.074 \pm 0.007$ & $0.821 \pm 0.010$ \\
\textbf{FLAME} & $\mathbf{0.064 \pm 0.004}$ & $\mathbf{0.791 \pm 0.016}$ \\
\bottomrule
\end{tabular}
\end{table}

\subsection{Cluster Recovery under Cross-Dataset Heterogeneity}
\label{app:exp-cluster}

\begin{figure}[h]
\centering
\includegraphics[width=0.85\linewidth]{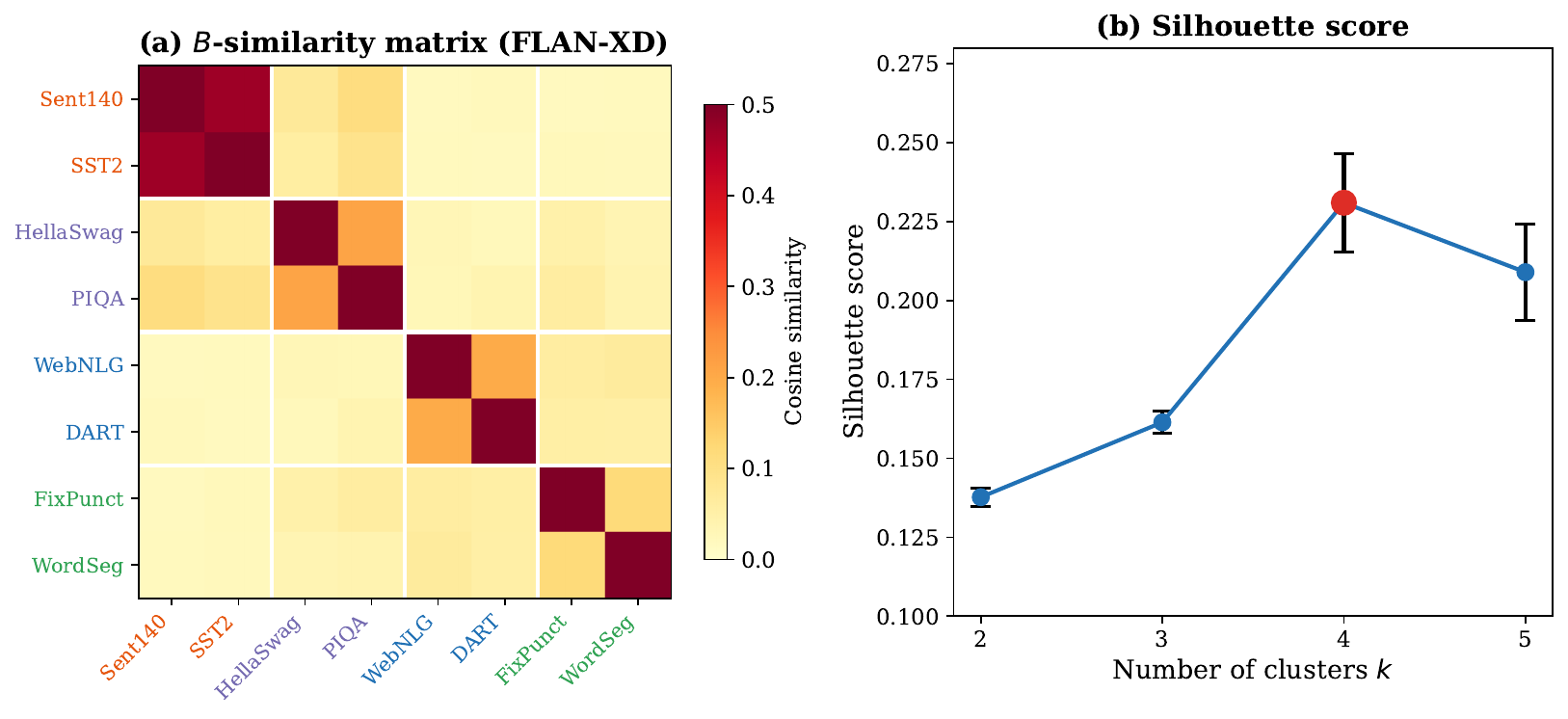}
\caption{Cluster recovery on FLAN-XD. \emph{(a)} $B$-similarity matrix \eqref{eq:cluster_dist}; block structure tracks the four-task partition, not the eight source datasets. \emph{(b)} Silhouette peaks at $k=4$.}
\label{fig:cluster_xd}
\end{figure}

We further test the server-side clustering of Section~\ref{sec:method-backbone} on FLAN-XD, where each task is represented by two distinct source datasets rather than IID replicas. Figure~\ref{fig:cluster_xd} shows that the entry-level $B$-similarity matrix exhibits clean $2\times 2$ task-pair blocks despite the surface-form shift between datasets within each pair, and the silhouette score peaks at $k=4$, recovering the underlying task partition exactly rather than splitting into eight dataset-specific singletons.

\section{Convergence Analysis of FLAME}
\label{sec:convergence}

We analyze the convergence of FLAME's clustered MoE training stage (Phase~3 of Section~\ref{sec:method-backbone}). The trainable individual expert is the only parameter group that receives a local gradient signal during this stage, and the per-step descent of its loss is what the bound below tracks. The Laplace stage of Section~\ref{sec:method-laplace} is a test-time procedure applied to converged MAP weights, so it lies outside the optimization scope; we return to it in Section~\ref{sec:conv-floor}, where the gate floor of the enlarged router is shown to supply the per-token Fisher mass it requires.

\subsection{Proof-Specific Setting}
\label{sec:conv-notation}

We adopt the notation of Section~\ref{sec:method} and introduce only quantities local to the proof.

\paragraph{Iteration index.} Phase~3 runs for $T$ rounds with $E$ local SGD steps per round, for $K=TE$ total steps. The global step is $k\in\{0,\ldots,K-1\}$ and its round is $t(k)\in\{1,\ldots,T\}$, with steps $(t-1)E,\ldots,tE-1$ belonging to round $t$.

\paragraph{Individual expert and frozen experts as Euclidean vectors.} The trainable individual factors are stacked across layers into $\theta_n^{(k)}=(A_n^{(k),(\ell)},B_n^{(k),(\ell)})_{\ell=1}^L\in\mathbb{R}^{P_{\rm ind}}$ with $P_{\rm ind}=Lr(d_{\rm in}+d_{\rm out})$, equipped with the Frobenius norm. The round-$t$ frozen-expert configuration $\bar{\mathbf{E}}^t=\{(\bar A^{(e),t},\bar B^{(e),t})\}_{e=1}^M$ is treated as a vector with the natural Frobenius norm.

\paragraph{Per-token gate on the individual expert.} The router output is $\phi_n(x)=G_n x$ with $G_n\in\mathbb{R}^{(2(M+1)-1)\times d_{\rm in}}$, mapped to experts by $\pi$ as in \eqref{eq:moe_forward}. The per-token gate weight on the individual expert at layer $\ell$ is
\begin{equation}
\label{eq:gate-individual}
g_{n,I}^{(\ell)}(x;G_n) := \sum_{i\in\mathrm{TopK}(\phi_n(x),M+1),\,\pi(i)=0} g_{n,i}^{(\ell)}(x;G_n).
\end{equation}

We will use repeatedly that $g_{n,I}^{(\ell)}$ depends on $x$ and $G_n$ but not on $\theta_n$, because $\phi_n(x)=G_n x$ does not involve $\theta_n$.

\paragraph{Local loss and SGD update.} Holding $\bar{\mathbf{E}}^t$ and $G_n$ fixed throughout round $t$, the per-sample loss is $F_n(\theta_n;\xi;\bar{\mathbf{E}}^t,G_n)$ and the local population loss is
$\mathcal{L}_n^t(\theta_n) := \mathbb{E}_{\xi\sim\mathcal{D}_n}[F_n(\theta_n;\xi;\bar{\mathbf{E}}^t,G_n)]$.
With learning rate $\eta>0$, the local SGD update is
\begin{equation}
\label{eq:sgd-flat-2}
\theta_n^{(k+1)} = \theta_n^{(k)} - \eta\,\nabla F_n(\theta_n^{(k)};\xi_n^{(k)};\bar{\mathbf{E}}^{t(k)},G_n),
\qquad \xi_n^{(k)}\sim\mathcal{D}_n ,
\end{equation}
with $\mathbb{E}[\nabla F_n(\theta_n^{(k)};\xi_n^{(k)})\mid\theta_n^{(k)}]=\nabla\mathcal{L}_n^{t(k)}(\theta_n^{(k)})$. The router $G_n$ is fixed within a round and absorbed into the loss notation throughout the analysis.

\begin{algorithm}[t]
\caption{FLAME: Federated Laplace-Aligned Mixture of Experts}
\label{alg:flame}
\begin{algorithmic}[1]
\Require Clients $\{1,\ldots,N\}$ with local data $\{\mathcal{D}_n\}_{n=1}^N$; frozen LLM weights $W^0$; LoRA rank $r$; number of pre-clustering rounds $T_w$; total rounds $T$; local epochs per round $E$; cluster-count upper bound $M_{\max}$; posterior samples $S$.
\Statex \textbf{Local warmup} \Comment{rounds $1,\ldots,T_w$}
\For{$t=1,\ldots,T_w$, each client $n$ \textbf{in parallel}}
    \State Train $(A_n,B_n)$ on $\mathcal{D}_n$ for $E$ epochs without aggregation.
\EndFor
\Statex \textbf{Clustering at the server}
\State Form the pairwise distance matrix \eqref{eq:cluster_dist} and partition clients into $\{\mathcal{C}_1,\ldots,\mathcal{C}_M\}$ by agglomerative clustering with $M\leq M_{\max}$ selected by silhouette score. 
\Statex \textbf{Clustered MoE training} \Comment{rounds $T_w{+}1,\ldots,T$}
\For{$t=T_w+1,\ldots,T$}
    \State \textbf{Server:} for each client $n\in \mathcal{C}_j$, build the intra-cluster RoW expert with $\mathcal{P}=\mathcal{C}_j\setminus\{n\}$ and the $M-1$ cross-cluster experts with $\mathcal{P}=\mathcal{C}_{j'}$ for $j'\neq j$, in each case using \eqref{eq:proj_avg}-\eqref{eq:coeff_avg} for $A$ and \eqref{eq:b_mean} for $B$.
    \State \textbf{Client $n$ in parallel:} load the frozen RoW and cross-cluster expert states; train $(A_n,B_n)$ and the router $\phi_n$ on $\mathcal{D}_n$ under \eqref{eq:moe_forward} for $E$ epochs; upload $(A_n,B_n)$ to the server.
\EndFor

\Statex \textbf{Post-hoc KFAC Laplace}
\Comment{per client, no communication}
\For{each client $n$ \textbf{in parallel}}
    \State \heng{Populate the KFAC factors
    $(\Omega_n^{F},\Lambda_n^{F})$ separately for
    $F\in\{A,B\}$ via \eqref{eq:kfac_factors},
    with the frozen experts and trained router fixed.}
    \State Select the prior precision $\alpha_n$
    by maximizing $\log p(\mathcal{D}_n;\alpha_n)$
    on a one-dimensional grid.
    \State \heng{Draw $S$ posterior factor pairs
    $(A_{n,s},B_{n,s})$ via \eqref{eq:posterior_sample}
    at every adapted layer and predict via
    \eqref{eq:bma}.}
\EndFor
\end{algorithmic}
\end{algorithm}

\subsection{Assumptions}
\label{sec:conv-assumptions}

\begin{assumption}[Per-layer core gradient]
\label{ass:phi}
There exist $L_\Phi,G_\Phi>0$ such that the $g_{n,I}$-stripped Jacobian factors $\Phi_n^{(\ell)}$ defined in~\eqref{eq:phi-def} below satisfy, for every $n,\xi,G_n,\bar{\mathbf{E}}$ visited during training and every $\theta,\theta'$,
\[
\text{(i)} \sum_{\ell=1}^L\|\Phi_n^{(\ell)}(\xi;\theta)-\Phi_n^{(\ell)}(\xi;\theta')\|^2\leq L_\Phi^2\|\theta-\theta'\|^2,
\quad
\text{(ii)}  \sum_{\ell=1}^L\|\Phi_n^{(\ell)}(\xi;\theta)\|^2\leq G_\Phi^2.
\]
\end{assumption}

\begin{assumption}[Bounded loss]
\label{ass:bounded}
There exists $\mathcal{L}_n^\star\in\mathbb{R}$ with $\mathcal{L}_n^t(\theta)\geq\mathcal{L}_n^\star$ for every $\theta$ and every $t$.
\end{assumption}

\begin{assumption}[Gate floor and ceiling]
\label{ass:gate}
There exists $g_{\min}\in(0,1]$ such that $g_{\min}\leq g_{n,I}^{(\ell)}(x;G_n)\leq 1$ for every $n,G_n,\ell$ and every $x$ in the support of $\mathcal{D}_n$.
\end{assumption}

The ceiling $\leq 1$ is automatic since $g_{n,I}^{(\ell)}$ is a sum of softmax weights. The floor $g_{\min}>0$ is structural for FLAME: the top-$(M+1)$ rule on $2(M+1)-1$ positions with $M+1$ wired to the individual expert forces, by pigeonhole, at least one individual position into every selection, and the post-softmax weight on the selected position is bounded away from $0$ on any compact range of router logits (preserved by standard regularization throughout training).

\begin{assumption}[Lipschitz in frozen experts]
\label{ass:freeze-smooth}
There exists $\mathcal{C}_S>0$ such that $|F_n(\theta;\xi;\bar{\mathbf{E}},G_n)-F_n(\theta;\xi;\bar{\mathbf{E}}',G_n)|\leq \mathcal{C}_S\|\bar{\mathbf{E}}-\bar{\mathbf{E}}'\|$ for every $n,\theta,\xi,G_n$ and every two configurations visited during training.
\end{assumption}

\begin{assumption}[Slowly-varying frozen experts]
\label{ass:slow}
Define $\delta_t:=\|\bar{\mathbf{E}}^{t+1}-\bar{\mathbf{E}}^t\|$, the cumulative drift $\Delta_T:=\sum_{t=1}^T\delta_t$ is finite.
\end{assumption}

\subsection{Per-step Lemmas}
\label{sec:conv-perstep}

\begin{lemma}[Per-layer Jacobian factorization]
\label{lem:jacobian}
For every $n,\xi=(x,y),\bar{\mathbf{E}},G_n$ and every adapted layer $\ell$, the per-layer gradient admits the exact factorization
\begin{equation}
\label{eq:jac-factor-2}
\nabla_{\theta_n^{(\ell)}}F_n(\theta_n;\xi;\bar{\mathbf{E}},G_n)
\;=\;g_{n,I}^{(\ell)}(x;G_n)\,\Phi_n^{(\ell)}(\xi;\theta_n;\bar{\mathbf{E}},G_n),
\end{equation}
where $\Phi_n^{(\ell)}$ is the per-layer Jacobian factor obtained by pulling the gate scalar $g_{n,I}^{(\ell)}$ out of the gradient.
\end{lemma}

\begin{proof}
By \eqref{eq:moe_forward}, the layer-$\ell$ output is
\begin{equation}
\label{eq:layer-output}
y^{(\ell)}_{\rm out}(x;\theta_n,\bar{\mathbf{E}},G_n)
\;=\;W^{0,(\ell)}x\;+\;g_{n,I}^{(\ell)}(x;G_n)\,B_n^{(\ell)}A_n^{(\ell)}\,x\;+\;S^{(\ell)}(x;\bar{\mathbf{E}},G_n),
\end{equation}
where $S^{(\ell)}(x;\bar{\mathbf{E}},G_n)=\sum_{e=1}^M g_{n,e}^{(\ell)}(x;G_n)\bar B^{(e),(\ell)}\bar A^{(e),(\ell)}x$ collects the frozen-expert contributions and depends on $x,\bar{\mathbf{E}},G_n$. 

Let $h_\ell:=\nabla_{y^{(\ell)}_{\rm out}}F_n$ denote the upstream gradient at layer $\ell$. Applying the chain rule to \eqref{eq:layer-output} and using that $g_{n,I}^{(\ell)}(x;G_n)$ does not depend on $\theta_n$, so it can be pulled out as a scalar,
\begin{align}
\nabla_{A_n^{(\ell)}}F_n &= g_{n,I}^{(\ell)}(x;G_n)\,(B_n^{(\ell)})^\top h_\ell\, x^\top, \label{eq:gradA}\\
\nabla_{B_n^{(\ell)}}F_n &= g_{n,I}^{(\ell)}(x;G_n)\,h_\ell\,(A_n^{(\ell)}x)^\top. \label{eq:gradB}
\end{align}
Define
\begin{equation}
\label{eq:phi-def}
\Phi_n^{(\ell)}(\xi;\theta_n;\bar{\mathbf{E}},G_n)
\;:=\;\bigl((B_n^{(\ell)})^\top h_\ell x^\top,\;h_\ell(A_n^{(\ell)}x)^\top\bigr);
\end{equation}
stacking \eqref{eq:gradA}--\eqref{eq:gradB} gives \eqref{eq:jac-factor-2}.
\end{proof}

\begin{remark}[No router-Jacobian remainder]
\label{rem:no-R}
Because $g_{n,I}^{(\ell)}(x;G_n)$ is independent of $\theta_n$, no remainder term arises from differentiating the gate weights, so \eqref{eq:jac-factor-2} is an exact equality. This holds because the router parameters $G_n$ and the LoRA factors $\theta_n$ are disjoint in FLAME.
\end{remark}

A direct consequence of Lemma~\ref{lem:jacobian} is the global identity
\begin{equation}
\label{eq:global-grad-norm}
\|\nabla F_n(\theta_n;\xi;\bar{\mathbf{E}},G_n)\|^2\;=\;\sum_{\ell=1}^L\bigl(g_{n,I}^{(\ell)}(x;G_n)\bigr)^2\bigl\|\Phi_n^{(\ell)}(\xi;\theta_n;\bar{\mathbf{E}},G_n)\bigr\|^2.
\end{equation}

\begin{lemma}[Effective $L_\Phi$-smoothness]
\label{lem:smoothness}
Under Assumptions~\ref{ass:phi} and~\ref{ass:gate}, $\mathcal{L}_n^t$ is $L_\Phi$-smooth for every fixed $t$.
\end{lemma}

\begin{proof}
By Lemma~\ref{lem:jacobian} and the $\theta$-independence of $g_{n,I}^{(\ell)}$,
\[
\|\nabla F_n(\theta;\xi)-\nabla F_n(\theta';\xi)\|^2
=\sum_\ell(g_{n,I}^{(\ell)})^2\|\Phi_n^{(\ell)}(\xi;\theta)-\Phi_n^{(\ell)}(\xi;\theta')\|^2
\leq L_\Phi^2\|\theta-\theta'\|^2,
\]
using the ceiling $g_{n,I}^{(\ell)}\leq 1$ and Assumption~\ref{ass:phi}(i). Jensen's inequality $\|\mathbb{E}Z\|\leq\mathbb{E}\|Z\|$ extends the bound to $\nabla\mathcal{L}_n^t$.
\end{proof}

\begin{lemma}[Bounded second moment]
\label{lem:variance}
Under Assumptions~\ref{ass:phi} and~\ref{ass:gate}, $\|\nabla F_n(\theta;\xi)\|^2\leq G_\Phi^2$ pointwise, hence $\mathbb{E}_\xi\|\nabla F_n(\theta;\xi)\|^2\leq G_\Phi^2$.
\end{lemma}

\begin{proof}
From \eqref{eq:global-grad-norm} and $g_{n,I}^{(\ell)}\leq 1$, $\|\nabla F_n(\theta;\xi)\|^2\leq\sum_\ell\|\Phi_n^{(\ell)}\|^2\leq G_\Phi^2$ by Assumption~\ref{ass:phi} (ii).
\end{proof}

\begin{lemma}[Per-step descent inequality]
\label{lem:descent}
Under Assumptions~\ref{ass:phi}, \ref{ass:bounded}, \ref{ass:gate}, and the SGD update~\eqref{eq:sgd-flat-2} with $\eta\in(0,1/L_\Phi]$, every step $k$ within round $t$ satisfies
\begin{equation}
\label{eq:descent-conclusion}
\mathbb{E}\!\left[\mathcal{L}_n^t(\theta_n^{(k+1)})\,\Big|\,\theta_n^{(k)}\right]
\;\leq\;\mathcal{L}_n^t(\theta_n^{(k)})\;-\;\frac{\eta}{2}\bigl\|\nabla\mathcal{L}_n^t(\theta_n^{(k)})\bigr\|^2\;+\;\frac{L_\Phi\eta^2}{2}G_\Phi^2.
\end{equation}
\end{lemma}

\begin{proof}
Lemma~\ref{lem:smoothness} and the standard descent lemma for $L_\Phi$-smooth functions give, deterministically,
\begin{equation}
\label{eq:descent-rawstep}
\mathcal{L}_n^t(\theta_n^{(k+1)})
\leq\mathcal{L}_n^t(\theta_n^{(k)}) - \eta\big\langle\nabla\mathcal{L}_n^t(\theta_n^{(k)}),\nabla F_n(\theta_n^{(k)};\xi_n^{(k)})\big\rangle + \tfrac{L_\Phi\eta^2}{2}\|\nabla F_n(\theta_n^{(k)};\xi_n^{(k)})\|^2.
\end{equation}
Take conditional expectation given $\theta_n^{(k)}$ on both sides. The cross term equals $\eta\|\nabla\mathcal{L}_n^t(\theta_n^{(k)})\|^2$ by unbiasedness. The quadratic term is bounded via $\mathbb{E}\|Z\|^2=\|\mathbb{E}Z\|^2+\mathrm{Var}(Z)$ with $\mathrm{Var}(Z)\leq\mathbb{E}\|Z\|^2\leq G_\Phi^2$ by Lemma~\ref{lem:variance}, giving $\leq\|\nabla\mathcal{L}_n^t\|^2+G_\Phi^2$. Substituting,
\[
\mathbb{E}[\mathcal{L}_n^t(\theta_n^{(k+1)})\mid\theta_n^{(k)}]
\leq\mathcal{L}_n^t(\theta_n^{(k)}) - \eta\bigl(1-\tfrac{L_\Phi\eta}{2}\bigr)\|\nabla\mathcal{L}_n^t(\theta_n^{(k)})\|^2 + \tfrac{L_\Phi\eta^2}{2}G_\Phi^2.
\]
Since $\eta\leq 1/L_\Phi\Rightarrow 1-L_\Phi\eta/2\geq 1/2$, this yields \eqref{eq:descent-conclusion}.
\end{proof}

\subsection{Frozen-Expert Drift Under Grassmannian Aggregation}
\label{sec:conv-drift}

We now bound the cumulative drift $\Delta_T$ of Theorem~\ref{thm:main} under FLAME's choice of aggregator. The Grassmannian construction \eqref{eq:proj_avg}--\eqref{eq:coeff_avg} can be rewritten as a projection of the Euclidean mean onto the consensus row-space (Equation~\eqref{eq:grass-as-projection} below). This rewriting separates the per-round change of $\bar A_\mathcal{P}$ into a basis-frame component, which the Grassmannian projection removes, and a row-space component, which both aggregators retain.

\paragraph{Projection form of the Grassmannian aggregate.} Let $\mathcal{P}\subseteq\{1,\ldots,N\}\setminus\{n\}$ be a peer set (intra-cluster RoW or cross-cluster), with consensus basis $\bar Q_\mathcal{P}^t$ from \eqref{eq:consensus} at round $t$. Define the consensus projector
\begin{equation}
\label{eq:Pi-def}
\Pi_\mathcal{P}^t \;:=\; (\bar Q_\mathcal{P}^t)^\top\bar Q_\mathcal{P}^t\;\in\;\mathbb{R}^{d_{\rm in}\times d_{\rm in}},
\end{equation}
which is a rank-$r$ orthogonal projector onto the consensus row-space. Combining \eqref{eq:coeff_avg} with the definition of $\bar A_\mathcal{P}^{\rm Eucl}=\frac{1}{|\mathcal P|}\sum_{m\in\mathcal P}A_m$ yields the explicit factorization
\begin{equation}
\label{eq:grass-as-projection}
\bar A_\mathcal{P}^{\rm Grass}(t) \;=\; \bar A_\mathcal{P}^{\rm Eucl}(t)\,\Pi_\mathcal{P}^t,
\end{equation}
i.e., the Grassmannian aggregator is the Euclidean mean projected onto the consensus row-space.

\begin{lemma}[Gauge invariance of the consensus projector]
\label{lem:gauge}
For any peer transformation $A_m\mapsto R_m A_m$ with $R_m\in\mathrm{GL}(r)$, the matrices $Q_m^\top Q_m$, $\bar P_\mathcal{P}$, and $\Pi_\mathcal{P}$ are invariant.
\end{lemma}

\begin{proof}
$\mathrm{rowspan}(R_m A_m)=\mathrm{rowspan}(A_m)$, so the orthogonal projector $Q_m^\top Q_m$ depends only on the row-space, not on the basis. The remaining quantities are constructed from $\{Q_m^\top Q_m\}_{m\in\mathcal{P}}$ via \eqref{eq:proj_avg} and \eqref{eq:Pi-def} and inherit gauge invariance.
\end{proof}

\begin{lemma}[Drift decomposition: Grass dominates Eucl]
\label{lem:grass-drift}
For any peer set $\mathcal{P}$ and consecutive rounds $t,t+1$,
\begin{equation}
\label{eq:grass-drift-bound}
\bigl\|\bar A_\mathcal{P}^{\rm Grass}(t+1)-\bar A_\mathcal{P}^{\rm Grass}(t)\bigr\|_F
\;\leq\;\bigl\|\bar A_\mathcal{P}^{\rm Eucl}(t+1)-\bar A_\mathcal{P}^{\rm Eucl}(t)\bigr\|_F\;+\;\bigl\|\bar A_\mathcal{P}^{\rm Eucl}(t)\bigr\|_F\,\bigl\|\Pi_\mathcal{P}^{t+1}-\Pi_\mathcal{P}^t\bigr\|_{\rm op}.
\end{equation}
In particular, when the consensus row-space is stable across rounds ($\Pi_\mathcal{P}^{t+1}=\Pi_\mathcal{P}^t$),
\begin{equation}
\label{eq:grass-drift-stable}
\bigl\|\bar A_\mathcal{P}^{\rm Grass}(t+1)-\bar A_\mathcal{P}^{\rm Grass}(t)\bigr\|_F
\;\leq\;\bigl\|\bar A_\mathcal{P}^{\rm Eucl}(t+1)-\bar A_\mathcal{P}^{\rm Eucl}(t)\bigr\|_F,
\end{equation}
with strict inequality whenever the Euclidean residual has a nonzero component outside the consensus row-space.
\end{lemma}

\begin{proof}
Using \eqref{eq:grass-as-projection}, add and subtract $\bar A_\mathcal{P}^{\rm Eucl}(t)\,\Pi_\mathcal{P}^{t+1}$:
\[
\bar A_\mathcal{P}^{\rm Grass}(t+1)-\bar A_\mathcal{P}^{\rm Grass}(t)
=\bigl(\bar A_\mathcal{P}^{\rm Eucl}(t+1)-\bar A_\mathcal{P}^{\rm Eucl}(t)\bigr)\Pi_\mathcal{P}^{t+1}+\bar A_\mathcal{P}^{\rm Eucl}(t)\bigl(\Pi_\mathcal{P}^{t+1}-\Pi_\mathcal{P}^t\bigr).
\]
Apply the triangle inequality $\|XY\|_F\leq\|X\|_F\|Y\|_{\rm op}$ and enforce $\|\Pi_\mathcal{P}^{t+1}\|_{\rm op}=1$ to the first term, we obtain \eqref{eq:grass-drift-bound}. When $\Pi_\mathcal{P}^{t+1}=\Pi_\mathcal{P}^t=\Pi$, the second term vanishes, and the bound reduces to \eqref{eq:grass-drift-stable}, with strict inequality iff the Euclidean residual has a nonzero component in $\Pi^\perp$. The Euclidean aggregator $\bar A_\mathcal{P}^{\rm Eucl}$ has no analogous projection step and incurs the residual in full.
\end{proof}

\paragraph{Consequence for $\Delta_T$ under FLAME's empirical regime.} Section~\ref{sec:asym} of the main text (cf.\ Figure~\ref{fig:observations}(a)) documents that within-cluster client $A$-row-spaces are tightly aligned across the converged training trajectory; equivalently, the consensus projectors $\Pi_\mathcal{P}^t$ change slowly across rounds, and the second term of \eqref{eq:grass-drift-bound} is dominated by the first component. By \eqref{eq:grass-drift-stable}, the per-round contribution of the $A$-factor to $\delta_t$ is strictly smaller under Grass than under Eucl whenever the Euclidean residual carries a basis-frame component outside the consensus row-space, i.e., the LoRA gauge ambiguity of Section~\ref{sec:asym}. The per-round contribution of the $B$-factor is identical under both aggregators (both use Euclidean averaging on $B$, by Observation~2). Hence
\begin{equation}
\label{eq:Delta-comparison}
\Delta_T^{\rm Grass}\;\leq\;\Delta_T^{\rm Eucl},
\end{equation}
with strict inequality whenever the LoRA gauge varies across rounds.

\begin{lemma}[Loss continuity across rounds]
\label{lem:freeze-loss}
Under Assumption ~\ref{ass:freeze-smooth}, for every $\theta$ and every two consecutive rounds $t,t+1$,
\begin{equation}
\label{eq:loss-jump}
\bigl|\mathcal{L}_n^{t+1}(\theta)-\mathcal{L}_n^t(\theta)\bigr|\;\leq\;\mathcal{C}_S\,\delta_t.
\end{equation}
\end{lemma}

\begin{proof}
Take expectation over $\xi\sim\mathcal{D}_n$ of the pointwise bound from Assumption~\ref{ass:freeze-smooth} with $\|\bar{\mathbf{E}}^{t+1}-\bar{\mathbf{E}}^t\|=\delta_t$, and apply $|\mathbb{E}[X-Y]|\leq\mathbb{E}|X-Y|$.
\end{proof}

\subsection{Main Convergence Theorem}
\label{sec:conv-main}

\begin{theorem}[Convergence of FLAME's individual expert]
\label{thm:main}
Under Assumptions~\ref{ass:phi}--\ref{ass:slow}, the SGD update~\eqref{eq:sgd-flat-2} runs for $K=TE$ steps with learning rate $\eta\in(0,1/L_\Phi]$, for every client $n$,
\begin{equation}
\label{eq:thm-main-2}
\frac{1}{K}\sum_{k=0}^{K-1}\mathbb{E}\bigl\|\nabla\mathcal{L}_n^{t(k)}(\theta_n^{(k)})\bigr\|^2
\;\leq\;
\underbrace{\frac{2(\mathcal{L}_n^1(\theta_n^{(0)})-\mathcal{L}_n^\star)}{\eta K}}_{\text{initial gap}}
+\underbrace{L_\Phi\,\eta\,G_\Phi^2}_{\text{noise floor}}
+\underbrace{\frac{2\mathcal{C}_S\Delta_T}{\eta K}}
_{\text{aggregation drift}}.
\end{equation}
With $D_0:=\mathcal{L}_n^1(\theta_n^{(0)})-\mathcal{L}_n^\star+\mathcal{C}_S\Delta_T$ and the choice
\begin{equation}
\label{eq:eta-choice}
\eta=\min\!\Bigl(\tfrac{1}{L_\Phi},\sqrt{\tfrac{2D_0}{L_\Phi G_\Phi^2 K}}\Bigr),
\end{equation}
\eqref{eq:thm-main-2} simplifies to
\begin{equation}
\label{eq:thm-rate}
\frac{1}{K}\sum_{k=0}^{K-1}\mathbb{E}\bigl\|\nabla\mathcal{L}_n^{t(k)}(\theta_n^{(k)})\bigr\|^2
\;\leq\;2\sqrt{\tfrac{2L_\Phi G_\Phi^2 D_0}{K}}+\tfrac{2L_\Phi D_0}{K}
\;=\;\mathcal{O}\!\bigl(K^{-1/2}\bigr).
\end{equation}
The constant $\mathcal{C}_S\Delta_T$ in \eqref{eq:thm-main-2} is strictly smaller under FLAME's Grassmannian-on-$A$ aggregator than under Euclidean-on-$A$ aggregation, by \eqref{eq:Delta-comparison}.
\end{theorem}

\begin{proof}
\textbf{Step 1: Within-round telescope.} Sum the per-step descent~\eqref{eq:descent-conclusion} over $k=(t-1)E,\ldots,tE-1$ and take total expectation:
\begin{equation}
\label{eq:thm-roundtelescope}
\frac{\eta}{2}\sum_{k=(t-1)E}^{tE-1}\mathbb{E}\bigl\|\nabla\mathcal{L}_n^t(\theta_n^{(k)})\bigr\|^2
\;\leq\;\mathbb{E}\bigl[\mathcal{L}_n^t(\theta_n^{((t-1)E)})\bigr]-\mathbb{E}\bigl[\mathcal{L}_n^t(\theta_n^{(tE)})\bigr]+\tfrac{EL_\Phi\eta^2}{2}G_\Phi^2.
\end{equation}

\textbf{Step 2: Cross-round bridge.} The iterate $\theta_n^{(tE)}$ is shared between the end of round $t$ and the start of round $t+1$, but the loss switches from $\mathcal{L}_n^t$ to $\mathcal{L}_n^{t+1}$. Lemma~\ref{lem:freeze-loss} gives
\begin{equation}
\label{eq:thm-bridge}
\mathbb{E}\bigl[\mathcal{L}_n^{t+1}(\theta_n^{(tE)})\bigr]\;\leq\;\mathbb{E}\bigl[\mathcal{L}_n^t(\theta_n^{(tE)})\bigr]+\mathcal{C}_S\delta_t.
\end{equation}

\textbf{Step 3: Cross-round telescope.} Sum \eqref{eq:thm-roundtelescope} over $t=1,\ldots,T$ and chain via \eqref{eq:thm-bridge}: at each round boundary, $-\mathbb{E}[\mathcal{L}_n^t(\theta_n^{(tE)})]$ is replaced by $-\mathbb{E}[\mathcal{L}_n^{t+1}(\theta_n^{(tE)})]+\mathcal{C}_S\delta_t$, then cancels the matching term that opens round $t+1$. Only the global endpoints survive:
\begin{equation}
\label{eq:thm-grandtelescope}
\frac{\eta}{2}\sum_{k=0}^{K-1}\mathbb{E}\bigl\|\nabla\mathcal{L}_n^{t(k)}(\theta_n^{(k)})\bigr\|^2
\leq\mathcal{L}_n^1(\theta_n^{(0)})-\mathbb{E}\bigl[\mathcal{L}_n^T(\theta_n^{(TE)})\bigr]+\sum_{t=1}^{T-1}\mathcal{C}_S\delta_t+\tfrac{KL_\Phi\eta^2}{2}G_\Phi^2.
\end{equation}

\textbf{Step 4: Lower bound and rearrange.} Apply Assumption~\ref{ass:bounded} ($\mathbb{E}[\mathcal{L}_n^T(\theta_n^{(TE)})]\geq\mathcal{L}_n^\star$), use $\sum_{t=1}^{T-1}\delta_t\leq\Delta_T$, and divide by $\eta K/2$ to obtain \eqref{eq:thm-main-2}.

\textbf{Step 5: Closed-form rate.} The unconstrained minimizer of $\frac{2D_0}{\eta K}+L_\Phi\eta G_\Phi^2$ is $\eta^\star=\sqrt{2D_0/(L_\Phi G_\Phi^2 K)}$, attaining $2\sqrt{2L_\Phi G_\Phi^2 D_0/K}$. We split into two cases for~\eqref{eq:eta-choice}:
\begin{itemize} 
\item If $\eta^\star\leq 1/L_\Phi$ (large $K$), $\eta=\eta^\star$ gives bound $2\sqrt{2L_\Phi G_\Phi^2 D_0/K}$; the second term in \eqref{eq:thm-rate} is non-negative slack.
\item If $\eta^\star>1/L_\Phi$ (small $K$, equivalent to $K<2L_\Phi D_0/G_\Phi^2$), $\eta=1/L_\Phi$ gives initial-gap term $2L_\Phi D_0/K$ and noise-floor term $G_\Phi^2$. The condition $K<2L_\Phi D_0/G_\Phi^2$ implies $G_\Phi^2<2\sqrt{2L_\Phi G_\Phi^2 D_0/K}$, so the bound in \eqref{eq:thm-rate} dominates.
\end{itemize}

\textbf{Step 6: Comparison to Euclidean aggregation.} The cumulative drift $\Delta_T$ enters \eqref{eq:thm-main-2} (and $D_0$) linearly through the additive term $\tfrac{2\mathcal{C}_S\Delta_T}{\eta K}$, so the bound is monotone in $\Delta_T$; \eqref{eq:Delta-comparison} gives $\Delta_T^{\rm Grass}\leq\Delta_T^{\rm Eucl}$ with strict inequality whenever the LoRA gauge varies across rounds. The Grassmannian-aggregator bound is therefore strictly tighter in this regime.

\subsection{Role of the Gate Floor: Fisher Mass}
\label{sec:conv-floor}

Theorem~\ref{thm:main} uses the gate ceiling but not the floor. The floor $g_{\min}>0$ enters elsewhere: it gives a uniform per-token lower bound on the gradient on the individual expert, and hence on the Fisher mass accumulated by the post-hoc Laplace stage of Section~\ref{sec:method-laplace}.

\begin{lemma}[Per-step gradient lower bound]
\label{lem:lowerbound}
Under Assumption~\ref{ass:gate}, for every $n,\theta,\xi=(x,y),G_n,\bar{\mathbf{E}}$ and every $\ell$,
\begin{equation}
\label{eq:lowerbound-perstep}
\bigl\|\nabla_{\theta_n^{(\ell)}}F_n(\theta;\xi;\bar{\mathbf{E}},G_n)\bigr\|^2\;\geq\;g_{\min}^2\,\bigl\|\Phi_n^{(\ell)}(\xi;\theta;\bar{\mathbf{E}},G_n)\bigr\|^2.
\end{equation}
\end{lemma}

\begin{proof}
Lemma~\ref{lem:jacobian} gives $\|\nabla_{\theta_n^{(\ell)}}F_n\|=g_{n,I}^{(\ell)}\|\Phi_n^{(\ell)}\|\geq g_{\min}\|\Phi_n^{(\ell)}\|$ by the floor; squaring yields \eqref{eq:lowerbound-perstep}.
\end{proof}

\begin{corollary}[Uniform Fisher contribution]
\label{cor:fisher}
The Kronecker-factored Fisher \eqref{eq:kfac_factors} accumulated by FLAME's post-hoc Laplace stage receives a contribution from every token $\xi$ in the support of $\mathcal{D}_n$ with multiplicative weight at least $g_{\min}^2$, uniformly in $\xi$.
\end{corollary}

\begin{proof}
The KFAC factors are gradient outer products; \eqref{eq:lowerbound-perstep} provides a uniform pointwise lower bound on each token's contribution.
\end{proof}

\paragraph{Why FLAME and not FedALT.} For FLAME's enlarged-gate router, $g_{\min}>0$ is structurally guaranteed: pigeonhole on the $2(M+1)-1$-dimensional softmax forces at least one individual position into every top-$(M+1)$ selection, with post-softmax weight bounded away from $0$ on the compact range of router logits. For FedALT's unconstrained two-expert mixer, no such bound holds: the gate weight on the individual expert can be driven arbitrarily small on tokens where the mixer prefers the Row expert, so $g_{\min}=0$ in general. Corollary~\ref{cor:fisher} therefore applies to FLAME but not to FedALT, formalizing the qualitative contrast of Section~\ref{sec:method-laplace}.

\paragraph{Comparison to prior personalized federated LoRA analyses.}
The convergence analyses in FedALT \citep{fedalt} and FedLEASE \citep{fedlease} establish a parameter-sequence stability bound of the form $\|Z_k^{t+1}-Z_k^t\|\to\mathcal{O}(\epsilon/(1-\beta))$ under per-client strong convexity and smoothness, where $\beta$ is the Lipschitz constant of the optimal-parameter map with respect to the frozen aggregate. Theorem~\ref{thm:main} works in the nonconvex SGD setting and provides an $\mathcal{O}(K^{-1/2})$ rate to a stationary point, so the two analyses target different quantities and are not directly comparable as numerical statements. What is specific to our bound is that each additive term in \eqref{eq:thm-main-2} traces back to a single FLAME component: the smoothness and variance constants $L_\Phi$ and $G_\Phi^2$ are induced by the gate ceiling of the enlarged router, the cumulative drift is controlled by the Grassmannian aggregator through Lemma~\ref{lem:grass-drift}, and the Fisher-mass condition used by the Laplace stage is supplied by the gate floor through Corollary~\ref{cor:fisher}. The FedALT and FedLEASE bounds depend on $\beta$ alone, which is consistent with neither method introducing a manifold-aware aggregator or a Bayesian post-processing step that would require this kind of decomposition.

\end{proof}


\begin{thebibliography}{41}
\providecommand{\natexlab}[1]{#1}
\providecommand{\url}[1]{\texttt{#1}}
\expandafter\ifx\csname urlstyle\endcsname\relax
  \providecommand{\doi}[1]{doi: #1}\else
  \providecommand{\doi}{doi: \begingroup \urlstyle{rm}\Url}\fi

\bibitem[Absil et~al.(2004)Absil, Mahony, and Sepulchre]{absil2004riemannian}
P-A Absil, Robert Mahony, and Rodolphe Sepulchre.
\newblock Riemannian geometry of grassmann manifolds with a view on algorithmic
  computation.
\newblock \emph{Acta Applicandae Mathematicae}, 2004.

\bibitem[Achituve et~al.(2021)Achituve, Shamsian, Navon, Chechik, and
  Fetaya]{achituve2021personalized}
Idan Achituve, Aviv Shamsian, Aviv Navon, Gal Chechik, and Ethan Fetaya.
\newblock Personalized federated learning with gaussian processes.
\newblock In \emph{Proc. of Adv. Neural Inf. Process. Syst.}, 2021.

\bibitem[Bai et~al.(2024)Bai, Chen, Qian, Yao, and Li]{bai2024flexlora}
Jiamu Bai, Daoyuan Chen, Bingchen Qian, Liuyi Yao, and Yaliang Li.
\newblock Federated fine-tuning of large language models under heterogeneous
  tasks and client resources.
\newblock In \emph{Proc. of Adv. Neural Inf. Process. Syst.}, 2024.

\bibitem[Bian et~al.(2026)Bian, Wang, Zhang, and Xu]{fedalt}
Jieming Bian, Lei Wang, Letian Zhang, and Jie Xu.
\newblock Fedalt: Federated fine-tuning through adaptive local training with
  rest-of-world lora.
\newblock In \emph{Proc. AAAI Conf. Artif. Intell.}, 2026.
\newblock arXiv:2503.11880.

\bibitem[Brown et~al.(2020)Brown, Mann, Ryder, Subbiah, Kaplan, Dhariwal,
  Neelakantan, Shyam, Sastry, Askell, et~al.]{brown2020language}
Tom Brown, Benjamin Mann, Nick Ryder, Melanie Subbiah, Jared~D Kaplan, Prafulla
  Dhariwal, Arvind Neelakantan, Pranav Shyam, Girish Sastry, Amanda Askell,
  et~al.
\newblock Language models are few-shot learners.
\newblock In \emph{Proc. of Adv. Neural Inf. Process. Syst.}, volume~33, pp.\
  1877--1901, 2020.

\bibitem[Collins et~al.(2021)Collins, Hassani, Mokhtari, and
  Shakkottai]{collins2021exploiting}
Liam Collins, Hamed Hassani, Aryan Mokhtari, and Sanjay Shakkottai.
\newblock Exploiting shared representations for personalized federated
  learning.
\newblock In \emph{Proc. Intl. Conf. Mach. Learn.}, 2021.

\bibitem[Cong et~al.(2024)Cong, Daheim, Shen, Cremers, Yokota, Khan, and
  M{\"o}llenhoff]{tian2024variational}
Bai Cong, Nico Daheim, Yuesong Shen, Daniel Cremers, Rio Yokota,
  Mohammad~Emtiyaz Khan, and Thomas M{\"o}llenhoff.
\newblock Variational low-rank adaptation using ivon.
\newblock \emph{arXiv preprint arXiv:2411.04421}, 2024.

\bibitem[Daxberger et~al.(2021)Daxberger, Kristiadi, Immer, Eschenhagen, Bauer,
  and Hennig]{daxberger2021laplace}
Erik Daxberger, Agustinus Kristiadi, Alexander Immer, Runa Eschenhagen,
  Matthias Bauer, and Philipp Hennig.
\newblock Laplace redux-effortless bayesian deep learning.
\newblock In \emph{Proc. of Adv. Neural Inf. Process. Syst.}, 2021.

\bibitem[Devlin et~al.(2019)Devlin, Chang, Lee, and Toutanova]{kenton2019bert}
Jacob Devlin, Ming-Wei Chang, Kenton Lee, and Kristina Toutanova.
\newblock Bert: Pre-training of deep bidirectional transformers for language
  understanding.
\newblock In \emph{Proc. of NAACL}, 2019.

\bibitem[Ding et~al.(2023)Ding, Qin, Yang, Wei, Yang, Su, Hu, Chen, Chan, Chen,
  et~al.]{ding2023parameter}
Ning Ding, Yujia Qin, Guang Yang, Fuchao Wei, Zonghan Yang, Yusheng Su,
  Shengding Hu, Yulin Chen, Chi-Min Chan, Weize Chen, et~al.
\newblock Parameter-efficient fine-tuning of large-scale pre-trained language
  models.
\newblock \emph{Nature Machine Intelligence}, 5\penalty0 (3):\penalty0
  220--235, 2023.

\bibitem[Dou et~al.(2024)Dou, Zhou, Liu, Gao, Shen, Xiong, Zhou, Wang, Xi, Fan,
  et~al.]{dou2024loramoe}
Shihan Dou, Enyu Zhou, Yan Liu, Songyang Gao, Wei Shen, Limao Xiong, Yuhao
  Zhou, Xiao Wang, Zhiheng Xi, Xiaoran Fan, et~al.
\newblock Loramoe: Alleviating world knowledge forgetting in large language
  models via moe-style plugin.
\newblock In \emph{Proc. of Annu. Meet. Assoc. Comput. Linguist.}, pp.\
  1932--1945, 2024.

\bibitem[Fallah et~al.(2020)Fallah, Mokhtari, and
  Ozdaglar]{fallah2020personalized}
Alireza Fallah, Aryan Mokhtari, and Asuman Ozdaglar.
\newblock Personalized federated learning with theoretical guarantees: A
  model-agnostic meta-learning approach.
\newblock In \emph{Proc. of Adv. Neural Inf. Process. Syst.}, 2020.

\bibitem[Fan et~al.(2023)Fan, Kang, Ma, Chen, Wei, Fan, and Yang]{fan2023fate}
Tao Fan, Yan Kang, Guoqiang Ma, Weijing Chen, Wenbin Wei, Lixin Fan, and Qiang
  Yang.
\newblock Fate-llm: A industrial grade federated learning framework for large
  language models.
\newblock \emph{arXiv preprint arXiv:2310.10049}, 2023.

\bibitem[Fu et~al.(2023)Fu, Yang, So, Lam, Bing, and
  Collier]{fu2023effectiveness}
Zihao Fu, Haoran Yang, Anthony Man-Cho So, Wai Lam, Lidong Bing, and Nigel
  Collier.
\newblock On the effectiveness of parameter-efficient fine-tuning.
\newblock In \emph{Proc. AAAI Conf. Artif. Intell.}, 2023.

\bibitem[Hadi et~al.(2023)Hadi, Qureshi, Shah, Irfan, Zafar, Shaikh, Akhtar,
  Wu, Mirjalili, et~al.]{hadi2023survey}
Muhammad~Usman Hadi, Rizwan Qureshi, Abbas Shah, Muhammad Irfan, Anas Zafar,
  Muhammad~Bilal Shaikh, Naveed Akhtar, Jia Wu, Seyedali Mirjalili, et~al.
\newblock A survey on large language models: Applications, challenges,
  limitations, and practical usage.
\newblock \emph{Authorea Preprints}, 2023.

\bibitem[Hu et~al.(2022)Hu, Shen, Wallis, Allen-Zhu, Li, Wang, Wang, and
  Chen]{hu2022lora}
Edward~J. Hu, Yelong Shen, Phillip Wallis, Zeyuan Allen-Zhu, Yuanzhi Li, Shean
  Wang, Lu~Wang, and Weizhu Chen.
\newblock {LoRA}: Low-rank adaptation of large language models.
\newblock In \emph{Proc. Intl. Conf. Learn. Repr.}, 2022.

\bibitem[Kassab \& Simeone(2022)Kassab and Simeone]{kassab2022federated}
Rahif Kassab and Osvaldo Simeone.
\newblock Federated generalized bayesian learning via distributed stein
  variational gradient descent.
\newblock \emph{IEEE Transactions on Signal Processing}, 2022.

\bibitem[Kuang et~al.(2024)Kuang, Qian, Li, Chen, Gao, Pan, Xie, Li, Ding, and
  Zhou]{kuang2024federatedscope}
Weirui Kuang, Bingchen Qian, Zitao Li, Daoyuan Chen, Dawei Gao, Xuchen Pan,
  Yuexiang Xie, Yaliang Li, Bolin Ding, and Jingren Zhou.
\newblock Federatedscope-llm: A comprehensive package for fine-tuning large
  language models in federated learning.
\newblock In \emph{Proc. of ACM SIGKDD Conf. Knowl. Discov. Data Min.}, 2024.

\bibitem[Li et~al.(2026)Li, Zhang, and Giannakis]{li2026low}
Bingcong Li, Yilang Zhang, and Georgios~B Giannakis.
\newblock Low-rank adaptation redux for large models.
\newblock \emph{arXiv preprint arXiv:2604.21905}, 2026.

\bibitem[Liu et~al.(2024)Liu, Wu, Zhao, Zhu, Xu, Tian, and
  Zheng]{liu2023moelora}
Q.~Liu, X.~Wu, X.~Zhao, Y.~Zhu, D.~Xu, F.~Tian, and Y.~Zheng.
\newblock When {MOE} meets {LLMs}: Parameter efficient fine-tuning for
  multi-task medical applications.
\newblock In \emph{Proceedings of the 47th International ACM SIGIR Conference
  on Research and Development in Information Retrieval}, pp.\  1104--1114,
  2024.
\newblock \doi{10.1145/3626772.3657722}.

\bibitem[MacKay(1992)]{mackay1992practical}
David~JC MacKay.
\newblock A practical bayesian framework for backpropagation networks.
\newblock \emph{Neural computation}, 4\penalty0 (3):\penalty0 448--472, 1992.

\bibitem[Martens \& Grosse(2015)Martens and Grosse]{martens2015kfac}
James Martens and Roger Grosse.
\newblock Optimizing neural networks with kronecker-factored approximate
  curvature.
\newblock In \emph{Proc. Intl. Conf. Mach. Learn.}, 2015.

\bibitem[McMahan et~al.(2017)McMahan, Moore, Ramage, Hampson, and
  y~Arcas]{mcmahan2017communication}
Brendan McMahan, Eider Moore, Daniel Ramage, Seth Hampson, and Blaise~Aguera
  y~Arcas.
\newblock Communication-efficient learning of deep networks from decentralized
  data.
\newblock In \emph{Proc. Intl. Conf. Artif. Intell. Stat.}, 2017.

\bibitem[Meng et~al.(2026)Meng, Tang, and Wong]{florg2026}
Chuiyang Meng, Ming Tang, and Vincent~WS Wong.
\newblock Florg: Federated fine-tuning with low-rank gram matrices and
  procrustes alignment.
\newblock \emph{arXiv preprint arXiv:2602.17095}, 2026.

\bibitem[Minaee et~al.(2024)Minaee, Mikolov, Nikzad, Chenaghlu, Socher,
  Amatriain, and Gao]{minaee2024large}
Shervin Minaee, Tomas Mikolov, Narjes Nikzad, Meysam Chenaghlu, Richard Socher,
  Xavier Amatriain, and Jianfeng Gao.
\newblock Large language models: A survey.
\newblock \emph{arXiv preprint arXiv:2402.06196}, 2024.

\bibitem[Ritter et~al.(2018)Ritter, Botev, and Barber]{ritter2018scalable}
Hippolyt Ritter, Aleksandar Botev, and David Barber.
\newblock A scalable laplace approximation for neural networks.
\newblock In \emph{Proc. Intl. Conf. Learn. Repr.}, 2018.

\bibitem[Sun et~al.(2024)Sun, Li, Li, and Ding]{sun2024improving}
Youbang Sun, Zitao Li, Yaliang Li, and Bolin Ding.
\newblock Improving lora in privacy-preserving federated learning.
\newblock 2024.

\bibitem[Tan et~al.(2022)Tan, Yu, Cui, and Yang]{tan2022towards}
Alysa~Ziying Tan, Han Yu, Lizhen Cui, and Qiang Yang.
\newblock Towards personalized federated learning.
\newblock \emph{IEEE Trans. Neural Netw. Learn. Syst.}, 2022.

\bibitem[Tian et~al.(2024)Tian, Shi, Guo, Li, and Xu]{tian2024hydra}
Chunlin Tian, Zhan Shi, Zhijiang Guo, Li~Li, and Chengzhong Xu.
\newblock Hydralora: An asymmetric lora architecture for efficient fine-tuning.
\newblock In \emph{Proc. of Adv. Neural Inf. Process. Syst.}, 2024.

\bibitem[Touvron et~al.(2023)Touvron, Lavril, Izacard, Martinet, Lachaux,
  Lacroix, Rozi{\`e}re, Goyal, Hambro, Azhar, et~al.]{touvron2023llama}
Hugo Touvron, Thibaut Lavril, Gautier Izacard, Xavier Martinet, Marie-Anne
  Lachaux, Timoth{\'e}e Lacroix, Baptiste Rozi{\`e}re, Naman Goyal, Eric
  Hambro, Faisal Azhar, et~al.
\newblock Llama: Open and efficient foundation language models.
\newblock \emph{arXiv preprint arXiv:2302.13971}, 2023.

\bibitem[Wang et~al.(2026)Wang, Bian, Zhang, and Xu]{fedlease}
Lei Wang, Jieming Bian, Letian Zhang, and Jie Xu.
\newblock Adaptive lora experts allocation and selection for federated
  fine-tuning.
\newblock In \emph{Proc. of Adv. Neural Inf. Process. Syst.}, 2026.
\newblock arXiv:2509.15087.

\bibitem[Wang et~al.(2023)Wang, Aitchison, and Rudolph]{wang2023lora}
Xi~Wang, Laurence Aitchison, and Maja Rudolph.
\newblock Lora ensembles for large language model fine-tuning.
\newblock \emph{arXiv preprint arXiv:2310.00035}, 2023.

\bibitem[Wang et~al.(2024{\natexlab{a}})Wang, Shi, Han, Metaxas, and
  Wang]{wang2024blob}
Yibin Wang, Haizhou Shi, Ligong Han, Dimitris Metaxas, and Hao Wang.
\newblock Blob: Bayesian low-rank adaptation by backpropagation for large
  language models.
\newblock In \emph{Proc. of Adv. Neural Inf. Process. Syst.},
  2024{\natexlab{a}}.

\bibitem[Wang et~al.(2024{\natexlab{b}})Wang, Shen, He, Sun, Wang, Lyu, and
  Li]{wang2024flora}
Ziyao Wang, Zheyu Shen, Yexiao He, Guoheng Sun, Hongyi Wang, Lingjuan Lyu, and
  Ang Li.
\newblock Flora: Federated fine-tuning large language models with heterogeneous
  low-rank adaptations.
\newblock In \emph{Proc. of Adv. Neural Inf. Process. Syst.},
  2024{\natexlab{b}}.

\bibitem[Wilson \& Izmailov(2020)Wilson and Izmailov]{wilson2020bayesian}
Andrew~G Wilson and Pavel Izmailov.
\newblock Bayesian deep learning and a probabilistic perspective of
  generalization.
\newblock \emph{Proc. of Adv. Neural Inf. Process. Syst.}, 33:\penalty0
  4697--4708, 2020.

\bibitem[Yang et~al.(2023)Yang, Robeyns, Wang, and Aitchison]{yang2023bayesian}
Adam~X Yang, Maxime Robeyns, Xi~Wang, and Laurence Aitchison.
\newblock Bayesian low-rank adaptation for large language models.
\newblock \emph{arXiv preprint arXiv:2308.13111}, 2023.

\bibitem[Yang et~al.(2024{\natexlab{a}})Yang, Robeyns, Wang, and
  Aitchison]{yang2024laplace}
Adam~X. Yang, Maxime Robeyns, Xi~Wang, and Laurence Aitchison.
\newblock Bayesian low-rank adaptation for large language models.
\newblock In \emph{International Conference on Learning Representations},
  2024{\natexlab{a}}.

\bibitem[Yang et~al.(2024{\natexlab{b}})Yang, Long, Shen, Jiang, and
  Blumenstein]{long2024dual}
Yiyuan Yang, Guodong Long, Tao Shen, Jing Jiang, and Michael Blumenstein.
\newblock Dual-personalizing adapter for federated foundation models.
\newblock In \emph{Proc. of Adv. Neural Inf. Process. Syst.},
  2024{\natexlab{b}}.

\bibitem[Ye et~al.(2024)Ye, Wang, Chai, Li, Li, Xu, Du, Wang, and
  Chen]{ye2024openfedllm}
Rui Ye, Wenhao Wang, Jingyi Chai, Dihan Li, Zexi Li, Yinda Xu, Yaxin Du,
  Yanfeng Wang, and Siheng Chen.
\newblock Openfedllm: Training large language models on decentralized private
  data via federated learning.
\newblock In \emph{Proc. of ACM SIGKDD Conf. Knowl. Discov. Data Min.}, 2024.

\bibitem[Zhang et~al.(2024)Zhang, Vahidian, Kuo, Li, Zhang, Yu, Wang, and
  Chen]{zhang2023building}
Jianyi Zhang, Saeed Vahidian, Martin Kuo, Chunyuan Li, Ruiyi Zhang, Tong Yu,
  Guoyin Wang, and Yiran Chen.
\newblock Towards building the federatedgpt: Federated instruction tuning.
\newblock In \emph{Proc. of Intl. Conf. Acoust. Speech Signal Process.}, 2024.

\bibitem[Zhang et~al.(2023)Zhang, Chen, Bukharin, Karampatziakis, He, Cheng,
  Chen, and Zhao]{zhang2023adalora}
Qingru Zhang, Minshuo Chen, Alexander Bukharin, Nikos Karampatziakis, Pengcheng
  He, Yu~Cheng, Weizhu Chen, and Tuo Zhao.
\newblock Adalora: Adaptive budget allocation for parameter-efficient
  fine-tuning.
\newblock In \emph{Proc. Intl. Conf. Learn. Repr.}, 2023.

\end{thebibliography}
\end{document}